\documentclass[10pt,twocolumn,letterpaper]{article}

\usepackage{iccv}
\usepackage{times}
\usepackage{epsfig}
\usepackage{graphicx}
\usepackage{amsmath}
\usepackage{bm}
\usepackage{amssymb}
\usepackage{xcolor}
\usepackage{subfig}
\usepackage{caption}
\usepackage{comment}

\usepackage[pagebackref=true,breaklinks=true,letterpaper=true,colorlinks,bookmarks=false]{hyperref}

\iccvfinalcopy 

\def\iccvPaperID{****} 

\DeclareMathOperator*{\argmax}{arg\,max}

\ificcvfinal\fi
\begin{document}

\title{Weakly- and Semi-Supervised Object Detection with \\ Expectation-Maximization Algorithm}

\author{Ziang Yan, Jian Liang, Weishen Pan, Jin Li\\
Tsinghua University\\
Beijing, China\\
{\tt\small \{yza15,liangjian12,pws15,lijin14\}@mails.tsinghua.edu.cn}
\and
Changshui Zhang\\
Tsinghua University\\
Beijing, China\\
{\tt\small  zcs@mail.tsinghua.edu.cn}
}

\maketitle

\begin{abstract}
   Object detection when provided image-level labels instead of instance-level labels (\ie, bounding boxes) during training is an important problem in computer vision, since large scale image datasets with instance-level labels are extremely costly to obtain. In this paper, we address this challenging problem by developing an Expectation-Maximization (EM) based object detection method using deep convolutional neural networks (CNNs). Our method is applicable to both the weakly-supervised and semi-supervised settings. Extensive experiments on PASCAL VOC 2007 benchmark show that (1) in the weakly supervised setting, our method provides significant detection performance improvement over current state-of-the-art methods, (2) having access to a small number of strongly (instance-level) annotated images, our method can almost match the performace of the fully supervised Fast RCNN. We share our source code at \url{https://github.com/ZiangYan/EM-WSD}
\end{abstract}


\section{Introduction}\label{sec:intro}

Object detection, which is a fundamental problem in computer vision, aims to localize spatial extents of all instances of a particular object category. The state-of-the-art object detection approaches typically train deep Convolutional Neural Networks (CNNs) \cite{lecun1998gradient} from large scale image datasets with instance-level labels (\ie, bounding boxes) \cite{girshick2015fast,He_2016_CVPR,liu2016ssd,Redmon_2016_CVPR,ren2015faster}. A key bottleneck of these approaches is that they require instance-level labels (strong labels) during training, which are extremely costly to obtain. Image-level labels (weak labels), in the form of binary image labels that indicate which object categories are present in an image, are far easier to collect than detailed instance-level labels, especially given the growth of tagged images on the Internet \cite{guillaumin2009tagprop}. However, it still remains a challenging problem to utilize these image-level labels when training object detectors. In this paper, we develop an EM based object detection method for training CNN based object detectors from image-level labels, either alone or in combination with some instance-level labels. 

\begin{figure}[t!]
\begin{center}
\includegraphics[width=0.95\linewidth]{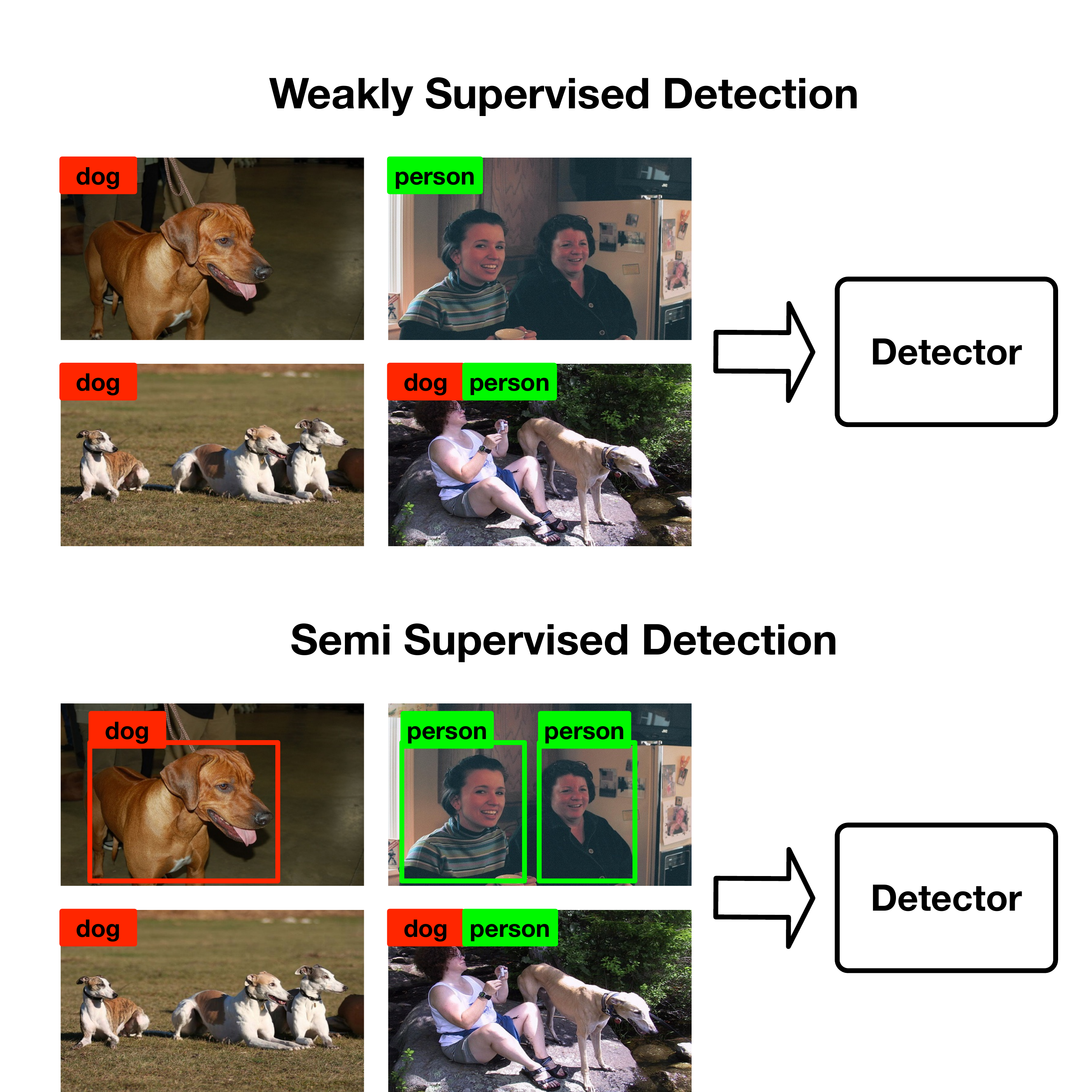}
\end{center}
   \caption{Weakly Supervised Detection and Semi Supervised Detection problem. In the training stage, only image-level labels are accessible in the weakly supervised setting, while image-level labels combined with some instance-level labels are available in the semi supervised setting. In the test stage, we predict bounding boxes for unseen images.}
\label{fig:pullfig}
\end{figure}

There are two different settings for training detectors given image-level labels: 1) learning detectors from image-level labels alone, and 2) learning detectors from image-level labels combined with some instance-level labels. In the first setting, the detection problem is also known as \emph{Weakly Supervised Detection}~(WSD), which has been explored in many recent literatures. Despite this progress, WSD is still far from solved, since the state-of-the-art performance of WSD on standard benchmarks \cite{Bilen_2016_CVPR,diba2016weakly,kantorov2016contextlocnet,yang2016weakly} is considerably lower than fully supervised counterparts \cite{girshick2015fast,ren2015faster}. In the second setting, the detection problem is known as \emph{Semi Supervised Detection}~(SSD). Several previous works \cite{hoffman2014lsda,Tang_2016_CVPR} assume the existence of several strongly annotated categories (all images in these categories have instance-level labels), and transfer knowledge from these strongly annotated categories to weakly annotated categories (all images in these categories only have image-level labels). We focus on a different setting, where both image-level labels and instance-level labels are in the same category, and additional strongly annotated categories are not required.

Although many existing WSD and SSD methods have shown promising results, there are three main drawbacks of existing approaches: (1) Existing WSD approaches do not cover the semi supervised setting, which is more practical in the real world application since SSD can achieve comparable detection performance to fully supervised methods with significant less instance-level labels. (2) Many WSD approaches treat object proposals in an image as independent instances, and invoke MI-SVM to mine positive proposals. They usually make a \emph{hard} decision for choosing a positive object proposal, thus support only one hypothesis at the same time. Intuitively, it's usually better to assign instance-level labels to proposals in a probabilistic fashion, especially when the detector is not sure which object proposal is positive. (3) Existing SSD methods usually transfer knowledge from auxiliary strongly annotated categories \cite{hoffman2014lsda,Tang_2016_CVPR,hoffman2015detector}. Tang \etal \cite{Tang_2016_CVPR} shows that visual and semantic similarities between weakly annotated categories and auxiliary strongly annotated categories play an essential role in improving the adaptation process. However, for a specific detection task, suitable additional strongly annotated categories are not always readily available.

In this paper, we present an EM based object detection method, which can be applied uniformly to both the semi supervised and weakly supervised settings. The EM algorithm estimates a probability distribution of missing values, thus it's smoother to optimize and can support multiple different hypotheses at the same time. This is important in the early training stage, since the estimation of instance-level labels for weakly annotated images is very noisy at that time. Given all observed data, we use the maximum likelihood estimation (MLE) to estimate the parameters of CNN (Section \ref{sec:objfun}). Treating instance-level labels as missing data for weakly annotated images, our method alternates between these two steps: 1) E-step: estimate a probability distribution over all possible latent locations, and 2) M-step: update the CNN weights using estimated locations from the last E-step (Section \ref{sec:emalgorithm}). In practice, the quality of the final extremum depends heavily on the initialization, since the whole optimization problem is highly non-convex. We use WSDDN by Bilen and Vedaldi \cite{Bilen_2016_CVPR} to initialize our EM algorithm (Section \ref{sec:integrate}). 

Our main contributions are:
\begin{itemize}
\item We present EM algorithms for object detection, applicable to both weakly supervised and semi supervised settings.
\item We show that our method outperforms the current state-of-the-art methods in the weakly supervised setting. On the PASCAL VOC 2007 test set, our method achieves 39.4\% mAP using AlexNet, 46.1\% mAP using VGG.
\item We show that by accessing a small number of strongly annotated images, our method can almost match the performace of the fully supervised detectors.
\end{itemize}

\section{Method}\label{sec:method}

\begin{figure*}[t]
\begin{center}
\vspace{-1em}\includegraphics[width=0.80\linewidth]{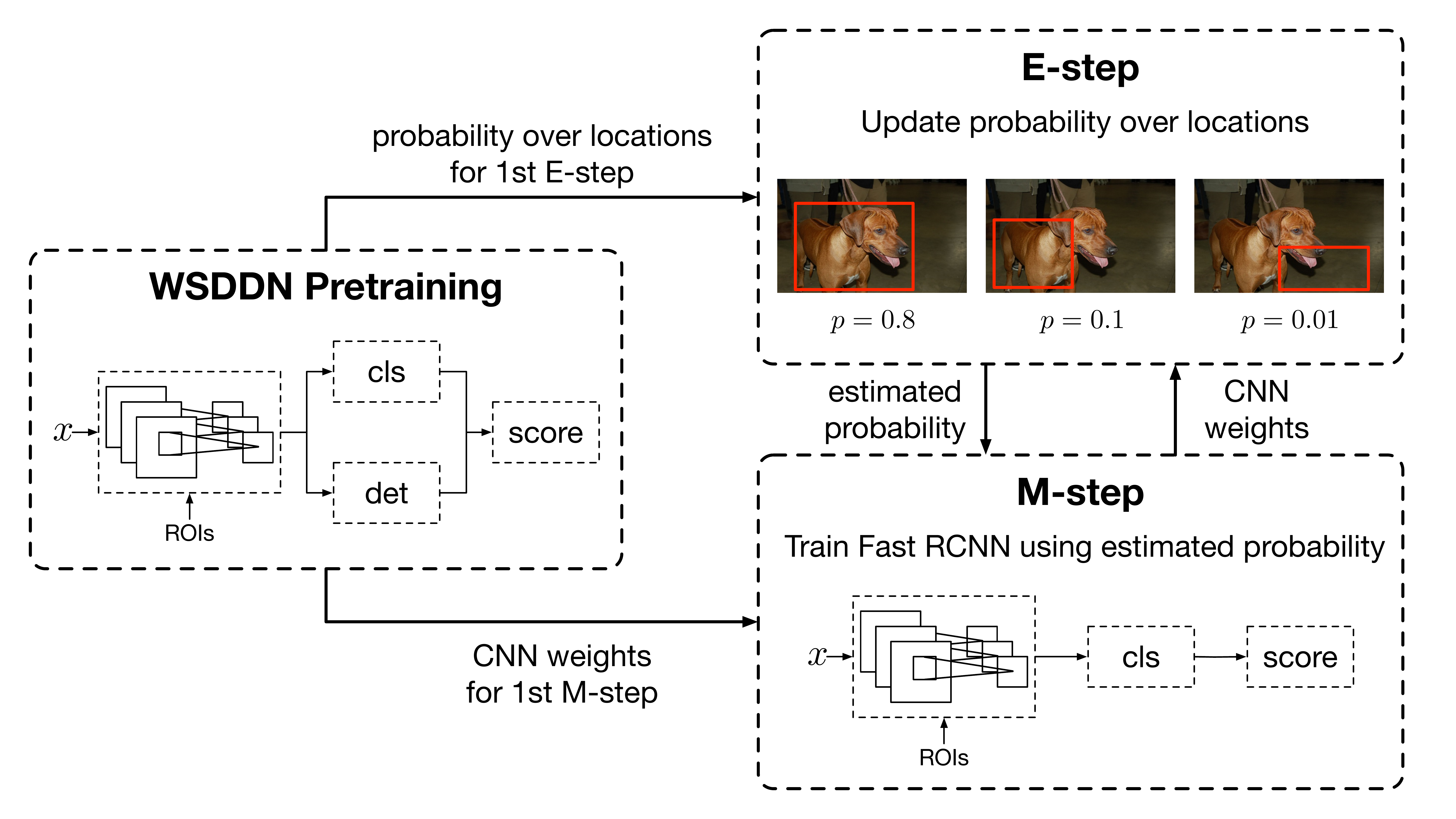}\vspace{-1.8em}
\end{center}
   \caption{Train object detector using the EM algorithm.}
\label{fig:pool}
\end{figure*}

In this section we introduce our EM based object detection method, which consists of a pre-training step followed by several EM iterations. Our method, which can not only learn from image-level labels, but also utilize possibly existing instance-level labels, is uniformly applicable to both SSD and WSD.

\vspace{.65em}

\noindent{\bf Notation.}\quad Let $\bm x\in\mathbb{R}^{H\times W\times 3}$ denote an image, where $H$ and $W$ are image height and width, respectively. An image $\bm x$ can be either weakly annotated or strongly annotated. We extract $B$ bounding box proposals from one image. In the fully supervised paradigm, each proposal is assigned to one of $C$ categories (including background category). Let $\bm{y}\in\{0,1\}^{B\times C}$ denote the instance-level label, where $\bm{y}_{ij}=1$ if the instance-level label of the $i$-th proposal is the $j$-th category. Let $\bm{z}\in\{0,1\}^{C}$ denote the image-level label, where $\bm{z}_{j}=1$ if the image contains the $j$-th object somewhere. Let $\bm\theta$ denote the vector of model parameters (\ie, the CNN weights). We denote by $\mathcal{S}$ the set of all pairs $(\bm x, \bm y)$, where $\bm x$ is a strongly annotated image and $\bm y$ is the instance-level label of $\bm x$. We denote by $\mathcal{W}$ the set of al pairs $(\bm x, \bm z)$, where $\bm x$ is a weakly annotated image and $\bm z$ is the image-level label of image $\bm x$.

\subsection{Objective function}\label{sec:objfun}
Using the maximum likelihood estimation, we maximize the joint likelihood for all observsed data, both weakly annotated and strongly annotated.

For strongly annotated images, the objective is to maximize $P(\bm y | \bm x;\bm\theta)$. As we train detectors in a region-based fashion, we maximize $P(\bm y | \bm x;\bm\theta)$ by maximizing the probability of each object proposal, which is similiar to \cite{girshick2014rich,girshick2015fast,ren2015faster}. We define the overall probability of an image as the product of probability of all object proposals
\begin{equation}
\prod_{(\bm x,\bm y)\in \mathcal{S}}P(\bm y|\bm x;\bm\theta)\triangleq\prod_{(\bm x,\bm y)\in \mathcal{S}}\prod_{i=1}^B P(\bm y_i|\bm b_i;\bm\theta),\label{eqn:factorize}
\end{equation}
where $\bm b_i$ is the $i$-th proposal of image $\bm x$, row vector $\bm y_i$ is the instance-level label of $\bm b_i$ (\ie, the $i$-th row of $\bm y$). 

For weakly annotated images, we treat $\bm y$ as missing data. We maximize $P(\bm z | \bm x;\bm\theta)$ for these images, since only the image-level label $\bm z$ is available for them. Note that once $\bm y$ is given, we can infer $\bm z$ deterministically by taking maximum over each column, thus $P(\bm z|\bm y)$ could be either $1$ or $0$. We denote by $\mathcal{Y}_{\bm z}$ the set of instance-level labels $\bm y$ that satisfy $P(\bm z|\bm y)=1$. Thus we have
\begin{align}
P(\bm z|\bm x;\bm\theta)&=\sum_{\text{all possible }\bm y}P(\bm y,\bm z|\bm x;\bm\theta)=\sum_{\bm y\in \mathcal{Y}_{\bm z}}P(\bm y|\bm x;\bm\theta).\label{eqn:allpossibley}
\end{align}

We find the maximum likelihood estimation of $\bm\theta$. Since we have observed both weakly annotated images $(\bm x, \bm z)\in\mathcal{W}$ and strongly annotated images $(\bm x, \bm y)\in\mathcal{S}$, the objective function (log likelihood for all observed data) is given by\footnote{Terms that do not depend on $\bm\theta$ are ignored.}
\begin{align}
J(\bm\theta)=\sum_{(\bm x,\bm y)\in \mathcal{S}}\log P(\bm y|\bm x;\bm\theta)+\sum_{(\bm x,\bm z)\in \mathcal{W}}\log P(\bm z|\bm x;\bm\theta),\label{eqn:objectivefun}
\end{align}
where $P(\bm y|\bm x;\bm\theta)$ and $P(\bm z|\bm x;\bm\theta)$ are given in \eqref{eqn:factorize} and \eqref{eqn:allpossibley}. We also apply $\ell 2$-norm regularization. Note the objective function \eqref{eqn:objectivefun} is uniformly applicable to both SSD and WSD, since the set $\mathcal{S}$ can be empty.

\subsection{EM algorithm}\label{sec:emalgorithm}

\noindent{\bf E-step.}\quad 
The purpose of the E-step is to estimate the complete-data log likelihood. For strongly annotated images, we have complete data $(\bm x,\bm y,\bm z)$, thus no estimation is required. For weakly annotated images, we estimate the complete-data log likelihood by taking expectation with respect to the latent variable $\bm y$.

Given the previously estimated parameter $\bm\theta'$, the expected complete-data log likelihood for weakly annotated image $\bm x$ and it's label $\bm z$ is given by\footnote{Details in supplementary material.}
\begin{align}
Q_{(\bm x,\bm z)}(\bm\theta;\bm\theta')
&=\sum_{\bm y\in \mathcal{Y}_z}P(\bm y|\bm x;\bm\theta')\log P(\bm y|\bm x;\bm\theta),\label{eqn:completedatalogweakly}
\end{align}
where $P(\bm y|\bm x;\bm\theta')$ and $P(\bm z|\bm y)$ are given in Sec \ref{sec:objfun}. The complete-data log likelihood for all images is
\begin{align}
Q(\bm\theta;\bm\theta')&=\sum_{(\bm x,\bm y)\in \mathcal{S}}\log P(\bm y|\bm x;\bm\theta)+\sum_{(\bm x,\bm z)\in \mathcal{W}}Q_{(\bm x,\bm z)}.\label{eqn:completedatalog}
\end{align}

\vspace{.65em}

\noindent{\bf M-step.}\quad 
In the M-step, we maximize $Q(\bm\theta;\bm\theta')$ with respect to $\bm\theta$. The key to maximize $Q(\bm\theta;\bm\theta')$ is maximizing $\log P(\bm y|\bm x;\bm\theta)$. Maximizing $\log P(\bm y|\bm x;\bm\theta)$ is a fully supervised detection problem, thus many region-based fully supervised detection approaches \cite{girshick2014rich,girshick2015fast,he2014spatial} can be used. We use Fast RCNN since it's simple but powerful.

\subsection{Integrate CNN into the EM algorithm}\label{sec:integrate}
We use a CNN to classify object proposals in the EM algorithm, since CNNs have excellent ability to learn visual features. The model parameter $\bm\theta$ we try to estimate, is the weights of the CNN. We use the same CNN architecture as in Fast RCNN: an ImageNet pre-trained CNN with a ROI pooling layer inserted in the middle, taking as input an entire image and a set of bounding box proposals. The network performs classification of the individual regions, by mapping each of them to a $C$-dimensional probability vector of class scores. The probability of an object proposal $P(\bm y_i|\bm b_i;\bm\theta)$ in \eqref{eqn:factorize} is given by the last softmax layer of CNN. In the M-step, we use SGD to optimize the expected complete-data log likelihood \eqref{eqn:completedatalog}. 

\vspace{.65em}

\noindent{\bf Spatial Consistence.}\quad Direct optimization of \eqref{eqn:completedatalog} is difficult, since there are too many terms to sum in $Q_{(\bm x,\bm z)}$ (\ie, the cardinality of $\mathcal{Y}_{\bm z}$ is too large). For example, if there are $M$ positive foreground categories in image $\bm x$, then the cardinality of $\mathcal{Y}_{\bm z}$ is $O\left((M+1)^B\right)$, which grows exponentially with respect to $B$. On the PASCAL VOC 2007, typically we have $B\sim 2000$, and $M=1,2,3$. We notice that \emph{Spatial Consistence} (SC) is powerful to reduce the number of terms in the sum in \eqref{eqn:completedatalogweakly}. SC refers to the fact that proposals usually contain the same object and share the same instance-level label, if they have large intersection over union (IoU) overlap. SC is used to sample foreground/background proposals in many region-based object detection approaches \cite{girshick2014rich,girshick2015fast,ren2015faster,Li_2016_CVPR,yang2016weakly}. In this paper, we use SC as a regularization technique for the latent space. We assume that there is only 1 object for each positive category, like many WSD approaches \cite{Li_2016_CVPR,yang2016weakly,Bilen_2016_CVPR}. We believe all \emph{reasonable} $\bm y$ can be generated by the following procedure. First, for each positive category $c$, choose 1 box as the center box and assign $c$ as the instance-level label to it. Second, assign $c$ as the instance-level label to all boxes that have at least 0.5 IoU with the center box, while assign the background category to other boxes. We simply ignore any $\bm y\in \mathcal{Y}_{\bm z}$ that can not be generated by the above procedure. Since there are about $B$ different choices in the first step, the number of all possible $\bm y$ reduces to $O\left(B^M\right)$.

\vspace{.65em}

\noindent{\bf Hard-EM.}\quad It's still tedious to sum $O\left(B^M\right)$ terms in \eqref{eqn:completedatalogweakly} if $M>1$. So we use two different approximation strategies: \emph{Hard-EM} and \emph{K-EM}. In Hard-EM, we keep only one term in \eqref{eqn:completedatalogweakly}
\begin{align}
Q_{(\bm x, \bm z)}(\bm\theta;\bm\theta')
&\approx P(\bm{y^*}|\bm x;\bm\theta')\log P(\bm{y^*}|\bm x;\bm\theta),\label{eqn:hardem}
\end{align}
where $\bm{y^*}$ is given by
\begin{align}
\bm{y^*}=\argmax_{\bm y\in \mathcal{Y}_{\bm z}}P(\bm y|\bm x;\bm\theta').
\end{align}

\vspace{.65em}

\noindent{\bf K-EM.}\quad Hard-EM may lose lots of information and hurt detection performance, since it discards too many terms in the summation in \eqref{eqn:completedatalogweakly}. K-EM achieves a better trade-off between information amount and computational cost, by keeping $K$ terms in \eqref{eqn:completedatalogweakly}. For each positive catetory $c$, we sort bounding boxes according to $P(\bm y_i=c|\bm b_i;\bm\theta')$ in descending order, and greedily keep the top $\sqrt[M]{K}$ bounding boxes. We set $K=100$ in all experiments.

\vspace{.65em}

\noindent{\bf Pre-training and post-processing.}\quad Learning CNN weights and localizing objects are two interconnected tasks. The whole optimization problem is highly non-convex, thus it's prone to get trapped into poor local extrema. In practice, the quality of the final extremum depends heavily on the initialization. In this paper we use WSDDN by Bilen and Vedaldi \cite{Bilen_2016_CVPR} to initialize our EM algorithm, since it can learn deep representation suitable for detection from only image-level labels. Investigating better initialization is left to future work.

We use the following strategy to transform WSDDN's two-stream network architecture to Fast RCNN's single-stream architecture. In the first E-step, we compute WSDDN's output score of \emph{center box} for each positive category in image $\bm x$. Then we set $P(\bm y|\bm x;\bm\theta')$ proportional to the product of these scores. In the first M-step, we initialize shared layers of Fast RCNN from WSDDN. Other reasonable transformation stragety can also be used.

Given a test image, we first generate around 2000 bounding box proposals using Edge Boxes \cite{zitnick2014edge}. Then, we score each proposal using our trained network. We perform the same post-processing as Fast RCNN: thresholding detected boxes class-by-class by their probabilities and then performing non maximum suppression with an overlap threshold of 0.4.

\section{Experiments}\label{sec:experiments}

\subsection{Dataset and evaluation metrics}

\noindent{\bf Dataset.}\quad We evaluate our method on the PASCAL VOC 2007 dataset, which is commonly used in object detection. The PASCAL VOC 2007 dataset consists of 2501 training images, 2510 validation images, and 5011 test images over 20 categories. We use both \emph{train} and \emph{val} splits as our training sets, and \emph{test} split as our test set.

\vspace{.65em}

\noindent{\bf Evaluation metrics.}\quad We use two metrics to evaluate detection performance. First, we evaluate detection mean Average Precision (mAP) on the PASCAL VOC 2007 \emph{test} split, following the standard PASCAL VOC protocol \cite{pascal-voc-2007}. Second, we compute CorLoc \cite{deselaers2012weakly} on the PASCAL VOC 2007 \emph{trainval} splits. CorLoc is the fraction of positive training images in which we localize an object of the target category correctly. Following \cite{pascal-voc-2007}, a detected bounding box is considered correct if it has at least 0.5 IoU with a ground truth bounding box.

\subsection{Experimental setup}
\noindent{\bf Network architectures.}\quad We use VGG16 \cite{Simonyan15} and AlexNet \cite{krizhevsky2012imagenet} as base CNN architectures for Fast RCNN. Pre-training on ImageNet classification data \cite{russakovsky2015imagenet} requires no bounding box annotations.

\vspace{.65em}

\noindent{\bf Data augmentation.}\quad Following WSDDN, we use multi-scale augmentation to achieve scale invariant object detection.  For AlexNet, we resize training images to six different scales (setting minimum of width or height to $\{400,600,750,880,1000,1200\}$). For VGG, we only use three different scales $\{400,600,900\}$ due to the limited GPU memory. We also apply horizontal flips to double the training set, as in Fast RCNN. Given a test image, we resize it to the same six scales, and each bounding box proposal is assigned to the scale such that the scaled bounding box is closest to $224^2$ pixels in area, as in SPPnet \cite{he2014spatial}.

\vspace{.65em}

\noindent{\bf Pretraining.}\quad We use the offical public implementation of WSDDN by Bilen and Vedaldi \cite{Bilen_2016_CVPR} in our pre-training step. In all experiments, we use the same hyperparameter configuration as \cite{Bilen_2016_CVPR}.

\vspace{.65em}

\noindent{\bf Training.}\quad  Following \cite{Li_2016_CVPR}, we generate around 2000 bounding box proposals for each image using Edge Boxes to train detectors. We use SGD to optimize the CNN weights. Every M-step consists of 40k SGD iterations. The learning rate is set to 0.001 in the first 30k iterations in the first M-step, and 0.0001 in all later iterations (\eg, we use 0.0001 learning rate in all 40k iterations in the second M-step). We finetune all layers after \emph{conv1} in all M-steps. We stop training after 3 M-steps (\ie., 120k SGD iterations in total). A momentum 0.9 and a weights decay of 0.0005 are used. The mini-batch is always constructed from 2 images: a randomly sampled image $\bm x$ and it's flipped image $\bm{x'}$. We sample 16 foreground proposals and 48 background proposals from each image, constructing a mini-batch of 128 proposals. Note that our sampling strategy differs from Fast RCNN, which sample the second image randomly rather than generate the horizontally flipped image from the first image. Note the mAP scores are typically 0.3 point worse if we use the sampling strategy in Fast RCNN. Using AlexNet, the whole training procedure takes about 10 hours with a Titan X Pascal GPU.

\vspace{.65em}

\noindent{\bf Reproducibility.}\quad Our implementation is based on the open-sourced Fast-RCNN code by Girshick \cite{girshick2015fast}, which is itself based on the excellent Caffe framework \cite{jia2014caffe}. We share our source code and the trained models at \url{https://github.com/ZiangYan/EM-WSD}.

\begin{table*}[t]
\begin{center}
\resizebox{\textwidth}{!}{\setlength\tabcolsep{3pt}%
\begin{tabular}{|l|cccccccccccccccccccc|c|}
\hline
Method & aero & bike & bird & boat & bottle & bus & car & cat & chair & cow & table & dog & horse & mbike & person & plant & sheep & sofa & train & tv & mAP\\
\hline
Cinbis \etal \cite{cinbis2014multi} & 35.8 & 40.6 & 8.1 & 7.6 & 3.1 & 35.9 & 41.8 & 16.8 & 1.4 & 23.0 & 4.9 & 14.1 & 31.9 & 41.9 & 19.3 & 11.1 & 27.6 & 12.1 & 31.0 & 40.6 & 22.4 \\
Song \etal \cite{song2014learning} & 27.6 & 41.9 & 19.7 & 9.1 & 10.4 & 35.8 & 39.1 & 33.6 & 0.6 & 20.9 & 10.0 & 27.7 & 29.4 & 39.2 & 9.1 & 19.3 & 20.5 & 17.1 & 35.6 & 7.1 & 22.7 \\
Bilen \etal \cite{bilen2014weakly} & 42.2 & 43.9 & 23.1 & 9.2 & 12.5 & 44.9 & 45.1 & 24.9 & 8.3 & 24.0 & 13.9 & 18.6 & 31.6 & 43.6 & 7.6 & 20.9 & 26.6 & 20.6 & 35.9 & 29.6 & 26.4\\
Li \etal, AlexNet \cite{Li_2016_CVPR} & 49.7 & 33.6 & 30.8 & 19.9 & 13 & 40.5 & 54.3 & 37.4 & 14.8 & 39.8 & 9.4 & 28.8 & 38.1 & 49.8 & 14.5 & 24.0 & 27.1 & 12.1 & 42.3 & 39.7 & 31.0 \\
Wang \etal \cite{wang2014weakly} & 48.9 & 42.3 & 26.1 & 11.3 & 11.9 & 41.3 & 40.9 & 34.7 & 10.8 & 34.7 & 18.8 & 34.4 & 35.4 & 52.7 & 19.1 & 17.4 & 35.9 & 33.3 & 34.8 & 46.5 & 31.6 \\
WSDDN\textsuperscript{\textdagger}, AlexNet \cite{Bilen_2016_CVPR} & 47.9 & 54.5 & 26.9 & 18.3 & 5.7 & 50.8 & 53.0 & 29.1 & 2.3 & 42.3 & 9.3 & 30.0 & 50.2 & 52.7 & 13.6 & 15.6 & 37.1 & 38.0 & 46.3 & 50.6 & 33.7 \\
ContextLocNet, VGG \cite{kantorov2016contextlocnet} & 57.1 & 52.0 & 31.5 & 7.6 & 11.5 & 55.0 & 53.1 & 34.1 & 1.7 & 33.1 & \bf{49.2} & 42.0 & 47.3 & 56.6 & 15.3 & 12.8 & 24.8 & 48.9 & 44.4 & 47.8 & 36.3 \\
WCCN, AlexNet \cite{diba2016weakly} & 43.9 & 57.6 & 34.9 & 21.3 & 14.7 & 64.7 & 52.8 & 34.2 & 6.5 & 41.2 & 20.5 & 33.8 & 47.6 & 56.8 & 12.7 & 18.8 & 39.6 & 46.9 & 52.9 & 45.1 & 37.3 \\
WSDDN-ENS, VGG \cite{Bilen_2016_CVPR} & 46.4 & 58.3 & 35.5 & 25.9 & 14.0 & 66.7 & 53.0 & 39.2 & 8.9 & 41.8 & 26.6 & 38.6 & 44.7 & 59.0 & 10.8 & 17.3 & 40.7 & 49.6 & 56.9 & 50.8 & 39.3 \\
Li \etal, VGG \cite{Li_2016_CVPR} & 54.5 & 47.4 & 41.3 & 20.8 & 17.7 & 51.9 & 63.5 & 46.1 & \bf{21.8} & \bf{57.1} & 22.1 & 34.4 & 50.5 & 61.8 & 16.2 & \bf{29.9} & 40.7 &15.9 & 55.3 & 40.2 & 39.5 \\
WCCN, VGG \cite{diba2016weakly} & 49.5 & 60.6 & 38.6 & \bf{29.2} & 16.2 & \bf{70.8} & 56.9 & 42.5 & 10.9 & 44.1 & 29.9 & 42.2 & 47.9 & 64.1 & 13.8 & 23.5 & 45.9 & \bf{54.1} & \bf{60.8} & \bf{54.5} & 42.8 \\
Ke \etal, VGG \cite{yang2016weakly} & 51.5 & \bf{66.1} & 45.5 & 19.4 & 11.0 & 56.6 & 64.5 & 57.3 & 3.0 & 51.1 & 42.7 & 41.8 & 51.9 & 64.8 & \bf{21.6} & 27.4 & 46.4 & 46.1 & 47.8 & 51.4 & 43.4 \\
\hline\hline
Hard-EM, no BB, AlexNet & 48.1 & 52.6 & 31.8 & 22.1 & 15.1 & 45.1 & 61.1 & 36.3 & 1.8 & 39.1 & 16.7 & 27.7 & 47.0 & 57.2 & 20.7 & 18.3 & 42.2 & 35.6 & 38.5 & 51.0 & 35.4 \\
Hard-EM, AlexNet & 58.3 & 59.0 & 35.7 & 21.8 & 15.2 & 50.9 & 64.5 & 39.2 & 2.3 & 47.3 & 14.8 & 34.3 & 52.8 & 60.6 & 13.0 & 18.9 & 44.9 & 39.8 & 37.9 & 51.7 & 38.1 \\
K-EM, AlexNet  & 56.6 & 60.9 & 34.3 & 24.0 & 19.0 & 54.1 & 64.8 & 41.6 & 6.1 & 47.0 & 18.3 & 24.2 & \bf{56.0} & 62.7 & 20.5 & 18.0 & 47.0 & 42.1 & 40.8 & 51.0 & 39.4 \\
K-EM, VGG & \bf{59.8} & 64.6 & \bf{47.8} & 28.8 & \bf{21.4} & 67.7 & \bf{70.3} & \bf{61.2} & 17.2 & 51.5 & 34.0 & \bf{42.3} & 48.8 & \bf{65.9} & 9.3 & 21.1 & \bf{53.6} & 51.4 & 54.7 & 50.7 & \bf{46.1} \\
\hline
\end{tabular}}
\end{center}
\caption{{\bf PASCAL VOC 2007 test} detection average precision (\%). In our experiments bounding box regression is applied by default. {\bf{no BB}}: train object detector without bounding box regression. \textsuperscript{\textdagger}\cite{Bilen_2016_CVPR} reports their results using VGG. For ablation studies we train a WSDDN initialized from AlexNet, and the source code is provided by the authors of \cite{Bilen_2016_CVPR}.}\label{table:mAPWeakly}
\end{table*}

\begin{table*}[t]
\begin{center}
\resizebox{\textwidth}{!}{\setlength\tabcolsep{3pt}%
\begin{tabular}{|l|cccccccccccccccccccc|c|}
\hline
Method & aero & bike & bird & boat & bottle & bus & car & cat & chair & cow & table & dog & horse & mbike & person & plant & sheep & sofa & train & tv & Avg\\
\hline
Siva \etal \cite{siva2012defence} & 45.8 & 21.8 & 30.9 & 20.4 & 5.3 & 37.6 & 40.8 & 51.6 & 7.0 & 29.8 & 27.5 & 41.3 & 41.8 & 47.3 & 24.1 & 12.2 & 28.1 &32.8 & 48.7 & 9.4 & 30.2 \\
Shi \etal \cite{shi2013bayesian} & 67.3 & 54.4 & 34.3 & 17.8 & 1.3 & 46.6 & 60.7 & 68.9 & 2.5 & 32.4 & 16.2 & 58.9 & 51.5 & 64.6 & 18.2 & 3.1 & 20.9 & 34.7 & 63.4 & 5.9 & 36.2 \\
Cinbis \etal \cite{cinbis2014multi} & 56.6 & 58.3 & 28.4 & 20.7 & 6.8 & 54.9 & 69.1 & 20.8 & 9.2 & 50.5 & 10.2 & 29.0 & 58.0 & 64.9 & 36.7 & 18.7 & 56.5 & 13.2 & 54.9 & 59.4 & 38.8 \\
Wang \etal \cite{wang2014weakly} & 80.1 & 63.9 & 51.5 & 14.9 & 21.0 & 55.7 & 74.2 & 43.5 & 26.2 & 53.4 & 16.3 & 56.7 & 58.3 & 69.5 & 14.1 & 38.3 & 58.8 & 47.2 & 49.1 & 60.9 & 48.5 \\
Li \etal, AlexNet \cite{Li_2016_CVPR} & 77.3 & 62.6 & 53.3 & 41.4 & 28.7 & 58.6 & 76.2 & 61.1 & 24.5 & 59.6 & 18.0 & 49.9 & 56.8 & 71.4 & 20.9 & 44.5 & 59.4 & 22.3 & 60.9 & 48.8 & 49.8 \\
Li \etal, VGG \cite{Li_2016_CVPR} & 78.2 & 67.1 & 61.8 & 38.1 & 36.1 & 61.8 & 78.8 & 55.2 & 28.5 & 68.8 & 18.5 & 49.2 & 64.1 & 73.5 & 21.4 & 47.4 & 64.6 &22.3 & 60.9 & 52.3 & 52.4 \\
WCCN, AlexNet \cite{diba2016weakly} & 79.7 & 68.1 & 60.4 & 38.9 & 36.8 & 61.1 & 78.6 & 56.7 & 27.8 & 67.7 & 20.3 & 48.1 & 63.9 & 75.1 & 21.5 & 46.9 & 64.8 & 23.4 & 60.2 & 52.4 & 52.6 \\
WSDDN\textsuperscript{\textdagger}, AlexNet \cite{Bilen_2016_CVPR} & 73.1 & 68.7 & 52.4 & 34.3 & 26.6 & 66.1 & 76.7 & 51.6 & 15.1 & 66.7 & 17.5 & 45.4 & 71.8 & 82.4 & 32.6 & 42.9 & 71.9 & 53.3 & 60.9 & 65.2 & 53.8 \\
ContextLocNet, VGG \cite{kantorov2016contextlocnet} & 83.3 & 68.6 & 54.7 & 23.4 & 18.3 & 73.6 & 74.1 & 54.1 & 8.6 & 65.1 & \bf{47.1} & \bf{59.5} & 67.0 & 83.5 & 35.3 & 39.9 & 67.0 & 49.7 & 63.5 & 65.2 & 55.1 \\
WCCN, VGG \cite{diba2016weakly} & \bf{83.9} & 72.8 & 64.5 & 44.1 & 40.1 & 65.7 & 82.5 & 58.9 & \bf{33.7} & 72.5 & 25.6 & 53.7 & 67.4 & 77.4 & 26.8 & 49.1 & 68.1 & 27.9 & 64.5 & 55.7 & 56.7 \\
WSDDN-ENS, VGG \cite{Bilen_2016_CVPR} & 68.9 & 68.7 & 65.2 & 42.5 & 40.6 & 72.6 & 75.2 & 53.7 & 29.7 & 68.1 & 33.5 & 45.6 & 65.9 & 86.1 & 27.5 & 44.9 & 76.0 & \bf{62.4} & 66.3 & 66.8 & 58.0 \\
\hline\hline
Hard-EM, no BB, AlexNet & 76.9 & 75.7 & 52.4 & 39.2 & 34.4 & 67.7 & 82.7 & 60.2 & 10.8 & 67.4 & 18.0 & 50.6 & 68.6 & 82.0 & 35.6 & 44.1 & 71.9 & 58.1 & 56.7 & 71.5 & 56.2 \\
Hard-EM, AlexNet & 81.1 & 77.0 & 56.7 & 40.3 & 31.6 & 72.6 & 85.8 & 62.9 & 14.8 & 74.5 & 17.5 & 54.9 & 76.0 & 85.3 & 34.0 & 46.1 & 76.0 & 54.1 & 63.6 & 71.1 & 58.8 \\
K-EM, AlexNet & 82.8 & 76.5 & 57.3 & 40.9 & 38.1 & 75.8 & 87.4 & 64.1 & 13.7 & 75.2 & 18.0 & 50.6 & \bf{77.7} & 85.3 & \bf{37.6} & 47.8 & 77.1 & 55.5 & 62.5 & 73.0 & 59.8 \\
K-EM, VGG & 79.8 & \bf{77.8} & \bf{66.7} & \bf{50.3} & \bf{57.0} & \bf{80.1} & \bf{89.9} & \bf{71.5} & 29.9 & \bf{75.9} & 30.5 & 58.9 & 73.2 & \bf{90.2} & 25.4 & \bf{51.8} & \bf{80.2} & 60.3 & \bf{72.4} & \bf{78.9} & \bf{65.0}\\
\hline
\end{tabular}}
\end{center}
\caption{{\bf PASCAL VOC 2007 trainval} correct localization (CorLoc \cite{deselaers2012weakly}) on positive images~(\%). {\bf no BB} and \textsuperscript{\textdagger}: see Table \ref{table:mAPWeakly}.}\label{table:CorLocWeakly}
\end{table*}
\subsection{Weakly supervised detection results}\label{sec:WSDresults}
\noindent{\bf Comparison with the state-of-the-art.}\quad We evaluate our method on the PASCAL VOC 2007 benchmark, using only image-level labels for training. In all experiments we set $K=100$, \ie, we compute around 100 different $\bm y$ for each weakly annotated image. We compare our results with the state-of-the-art methods for weakly supervised object detection in Table \ref{table:mAPWeakly}, \ref{table:CorLocWeakly}. 

In Table \ref{table:mAPWeakly} we report detection average precision (AP) on the PASCAL VOC 2007 \emph{test} set. Our best model, K-EM using VGG, achieves 46.1\% mAP and outperforms all recent state-of-the-art weakly supervised detection methods. Many previous works \cite{Li_2016_CVPR,diba2016weakly,song2014weakly,song2014learning,bilen2014weakly,wang2014weakly} use AlexNet as feature extractor. Cinbis \etal \cite{cinbis2014multi} also use Fisher Vector \cite{perronnin2010improving} and Edge Boxes objectness score \cite{zitnick2014edge}. For fair comparison we also train a model (K-EM, AlexNet in Table \ref{table:mAPWeakly}) using AlexNet in both pre-training and training. Our method achieves 39.4\% using AlexNet, outperforming the second best method \cite{diba2016weakly} using the same CNN architecture (39.4\% vs. 37.3\%).

In Table \ref{table:CorLocWeakly} we report correct localization (CorLoc) on the PASCAL VOC 2007 \emph{trainval} set. Our best model, K-EM using VGG, achieves 65.0\% average CorLoc for the 20 categories, outperforming the second best method \cite{Bilen_2016_CVPR} by 7 points (65.0\% vs. 58.0\%). Using AlexNet, our method achieves average CorLoc of 59.8\%, and outperform the current state-of-the-art method \cite{Bilen_2016_CVPR} using the same network by 6 points. We would like to highlight that even using AlexNet, our method also outperforms the all methods that use VGG as their base CNN architecture.

\begin{figure}[t]
\centering
\subfloat{\includegraphics[trim={0.5 12.5cm 7cm 0},clip,width=0.23\textwidth]{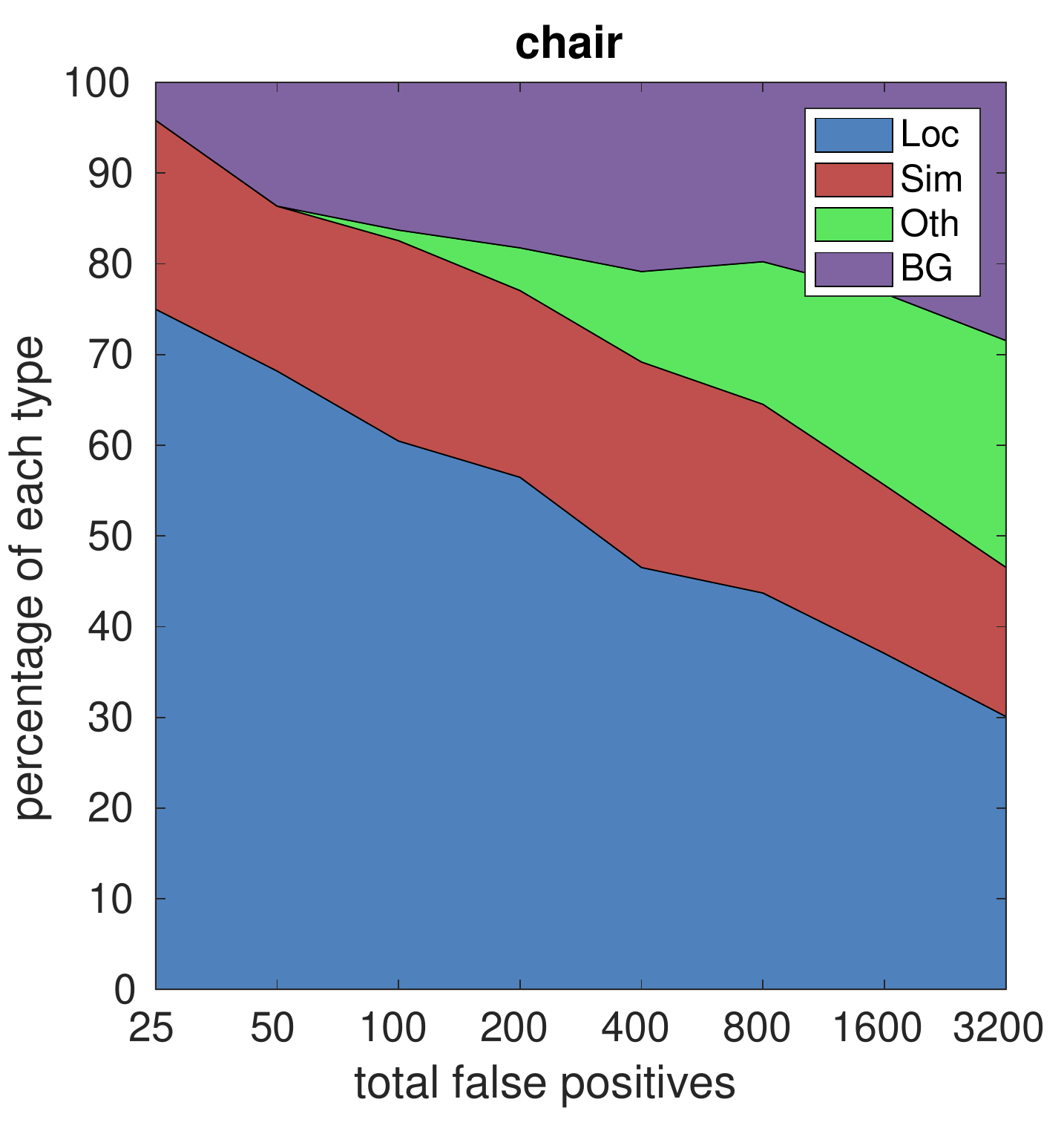}}%
\subfloat{\includegraphics[trim={0.5 12.5cm 7cm 0},clip,width=0.23\textwidth]{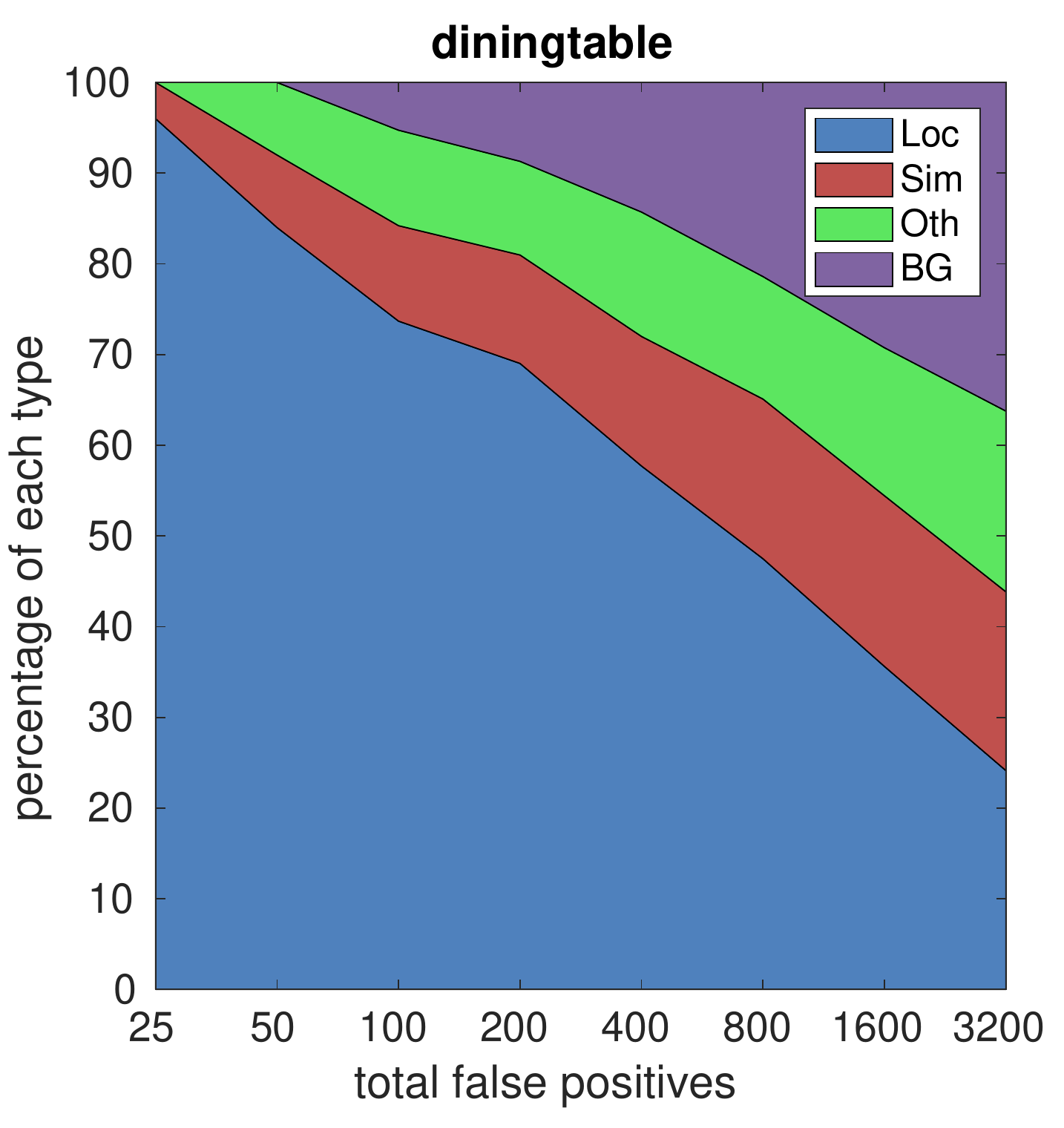}}\\%
\subfloat{\includegraphics[trim={0.5 12.5cm 7cm 0},clip,width=0.23\textwidth]{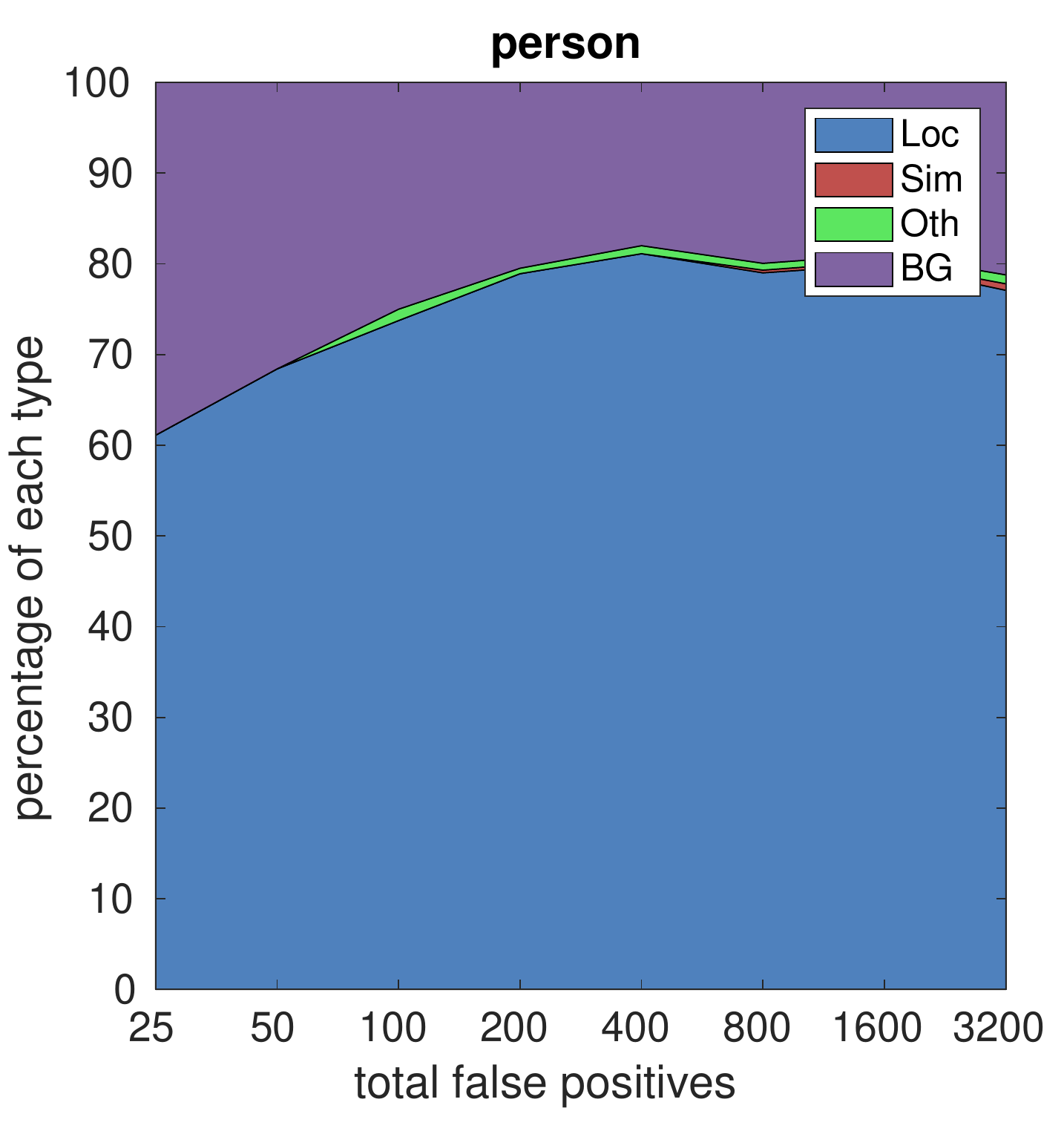}}%
\subfloat{\includegraphics[trim={0.5 12.5cm 7cm 0},clip,width=0.23\textwidth]{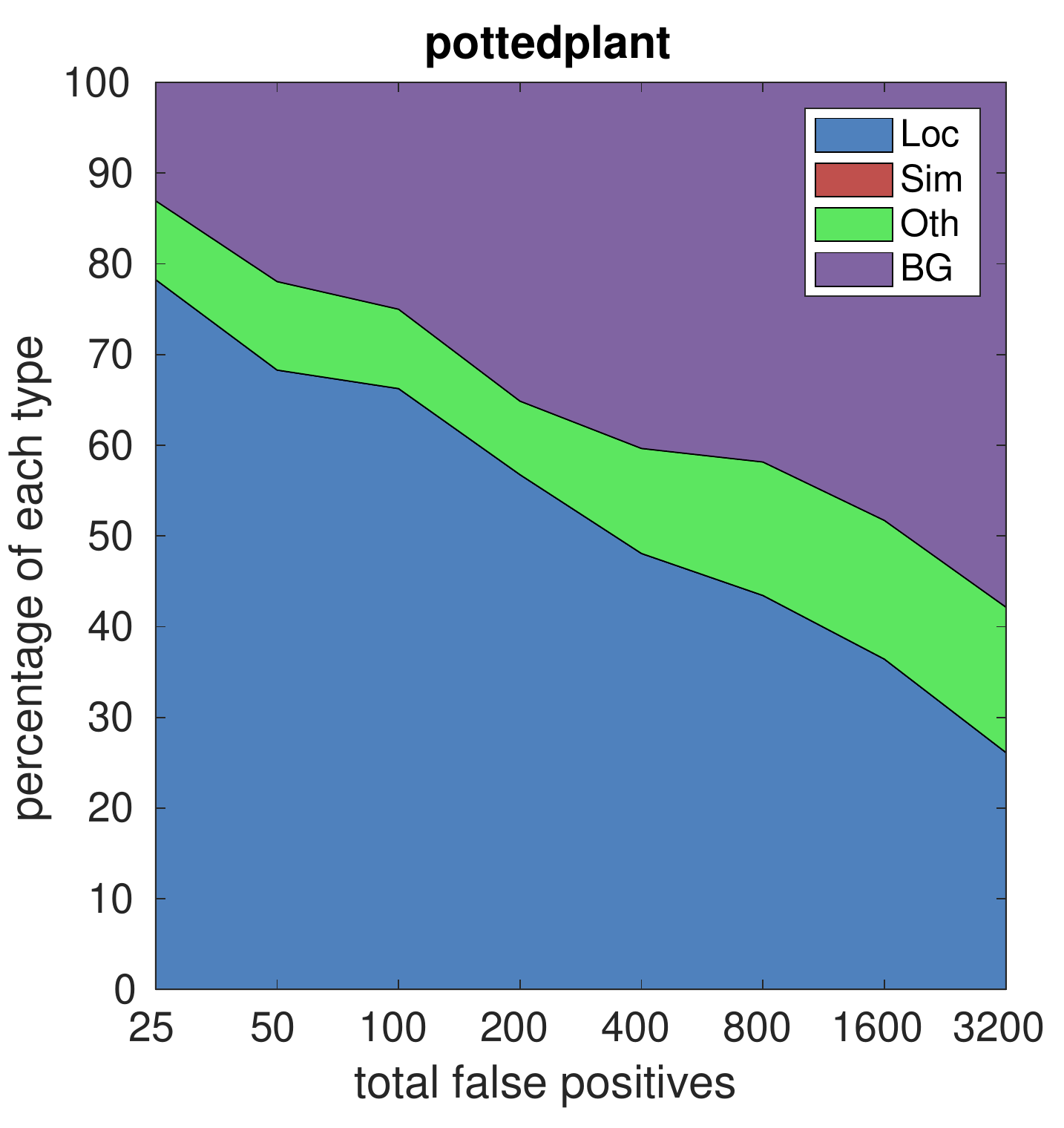}}%
\caption{Top-ranked false positive (FP) using K-EM + VGG in 4 object categories: chair, diningtable, person and pottedplant. Each false positive is categorized into 1 of 4 types: {\bf{Loc}}: poor localization, {\bf{Sim}}: confusion with a similar category, {\bf{Oth}}: confusion with other object category, {\bf{BG}}: confusion with background. Best viewed in color.}
\label{fig:error}
\end{figure}

\begin{figure*}[h!]
\begin{center}
\subfloat{\includegraphics[width=0.18\textwidth,height=0.10\textwidth]{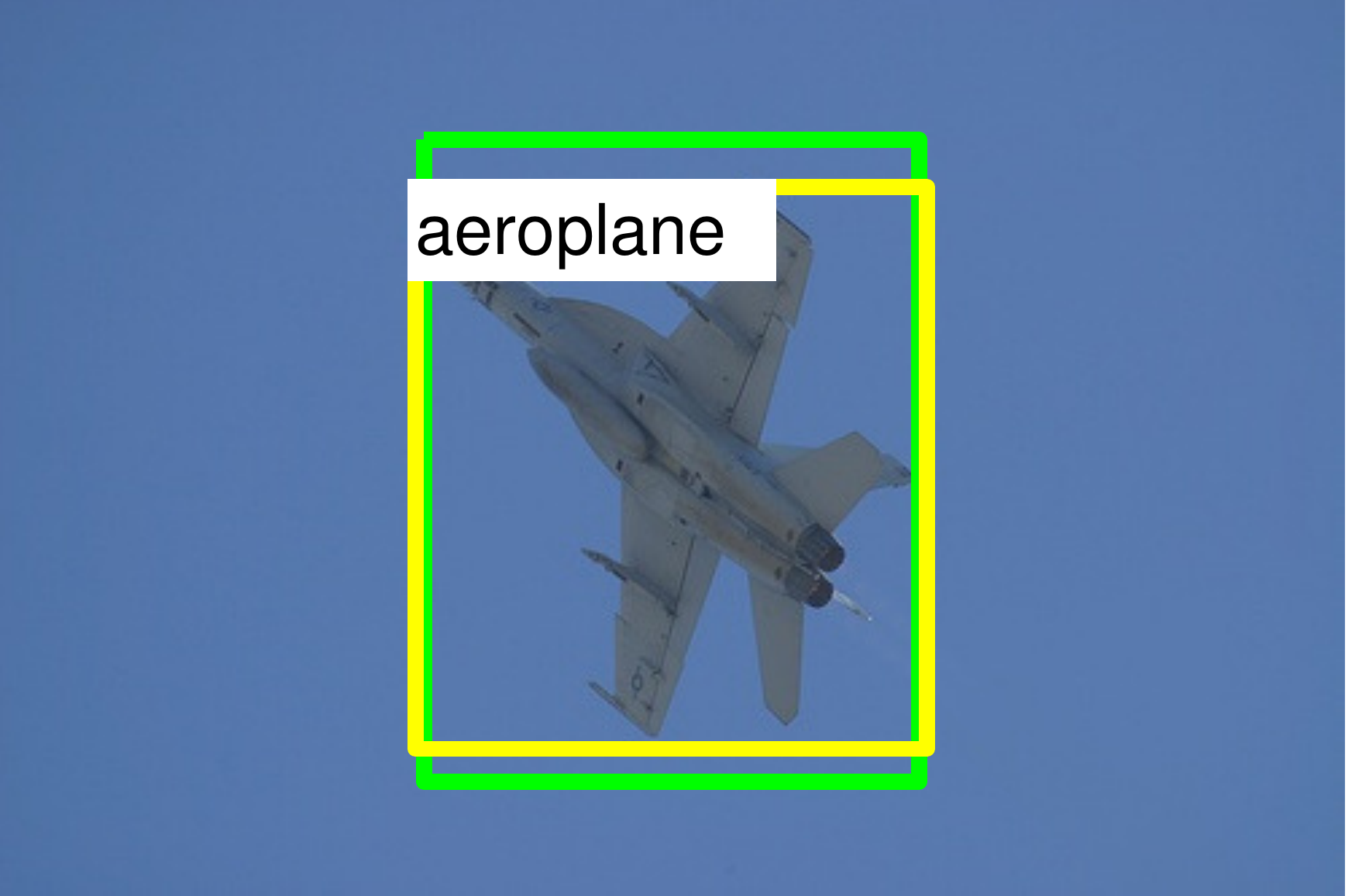}}\hspace{.3em}%
\subfloat{\includegraphics[width=0.18\textwidth,height=0.10\textwidth]{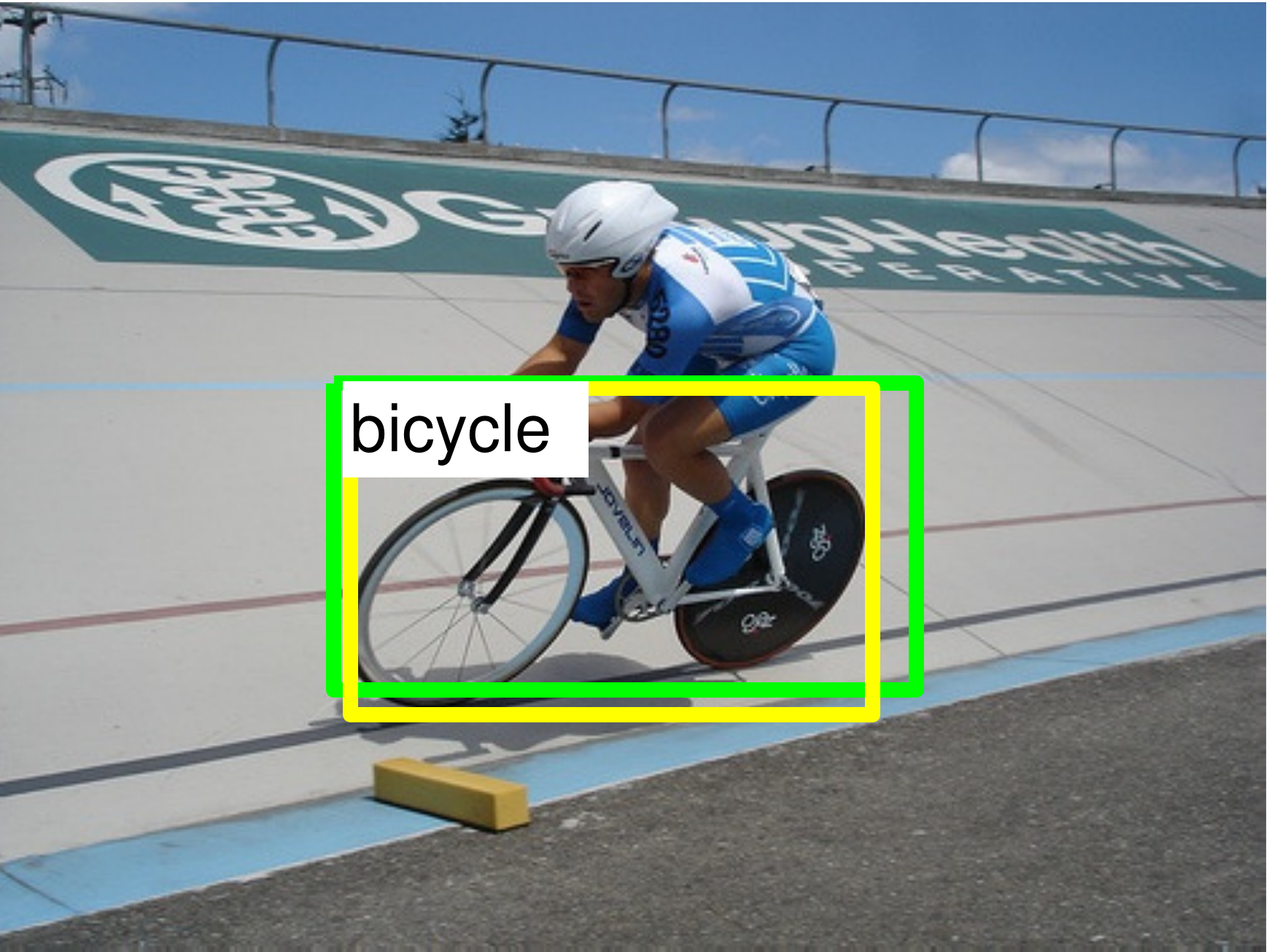}}\hspace{.3em}%
\subfloat{\includegraphics[width=0.18\textwidth,height=0.10\textwidth]{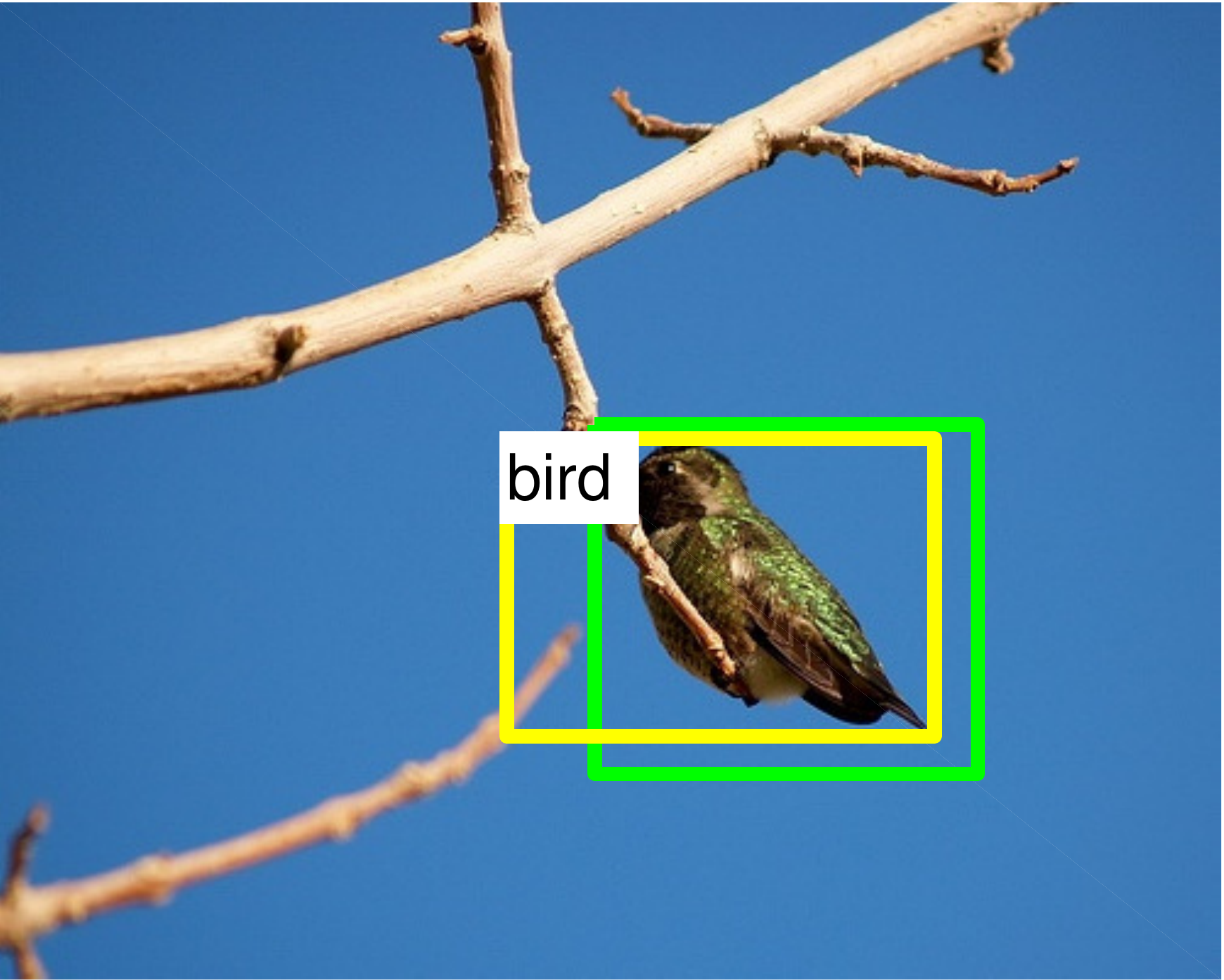}}\hspace{.3em}%
\subfloat{\includegraphics[width=0.18\textwidth,height=0.10\textwidth]{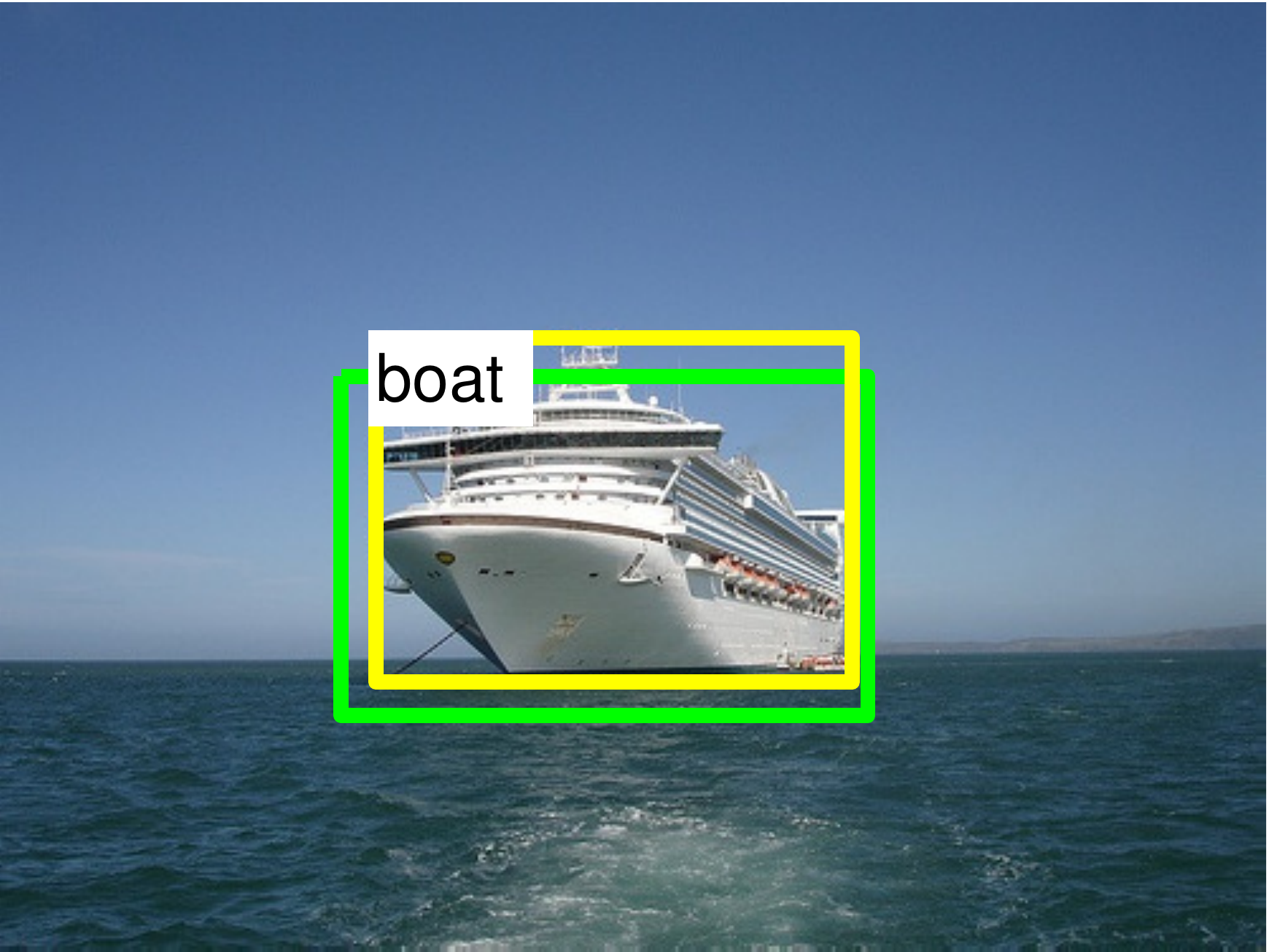}}\hspace{.3em}%
\subfloat{\includegraphics[width=0.18\textwidth,height=0.10\textwidth]{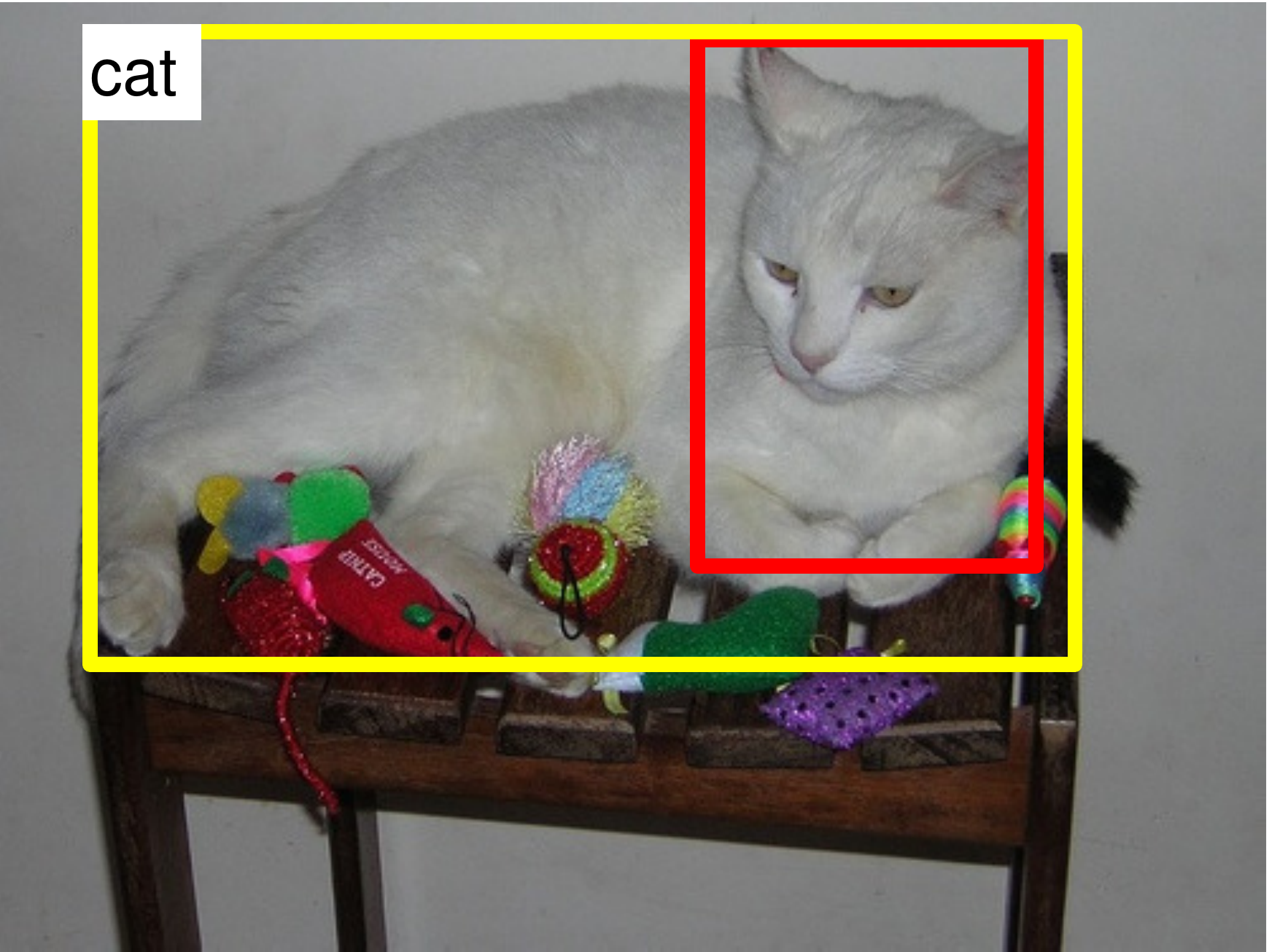}}\vspace{-.75em}\\%
\subfloat{\includegraphics[width=0.18\textwidth,height=0.10\textwidth]{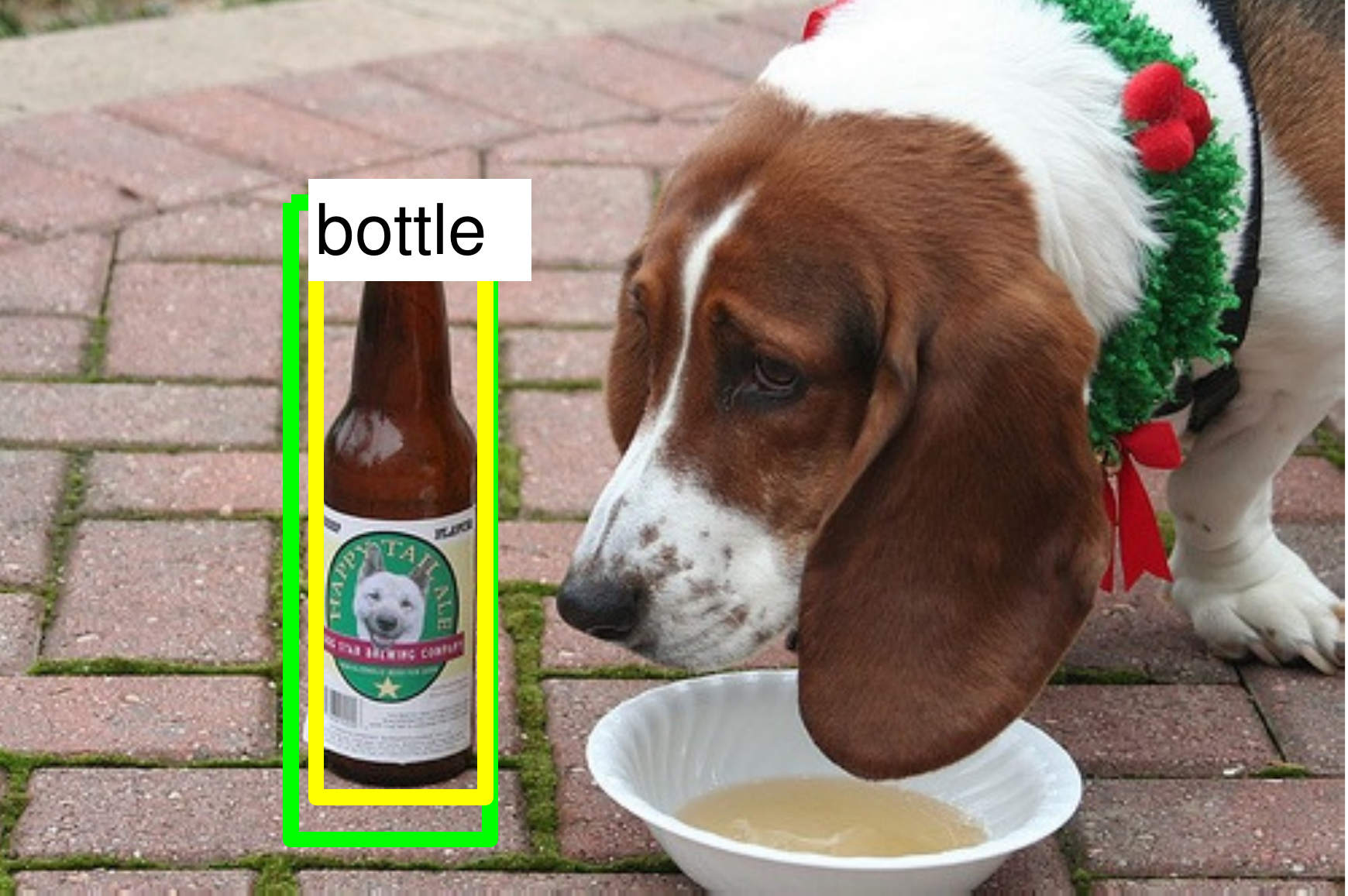}}\hspace{.3em}%
\subfloat{\includegraphics[width=0.18\textwidth,height=0.10\textwidth]{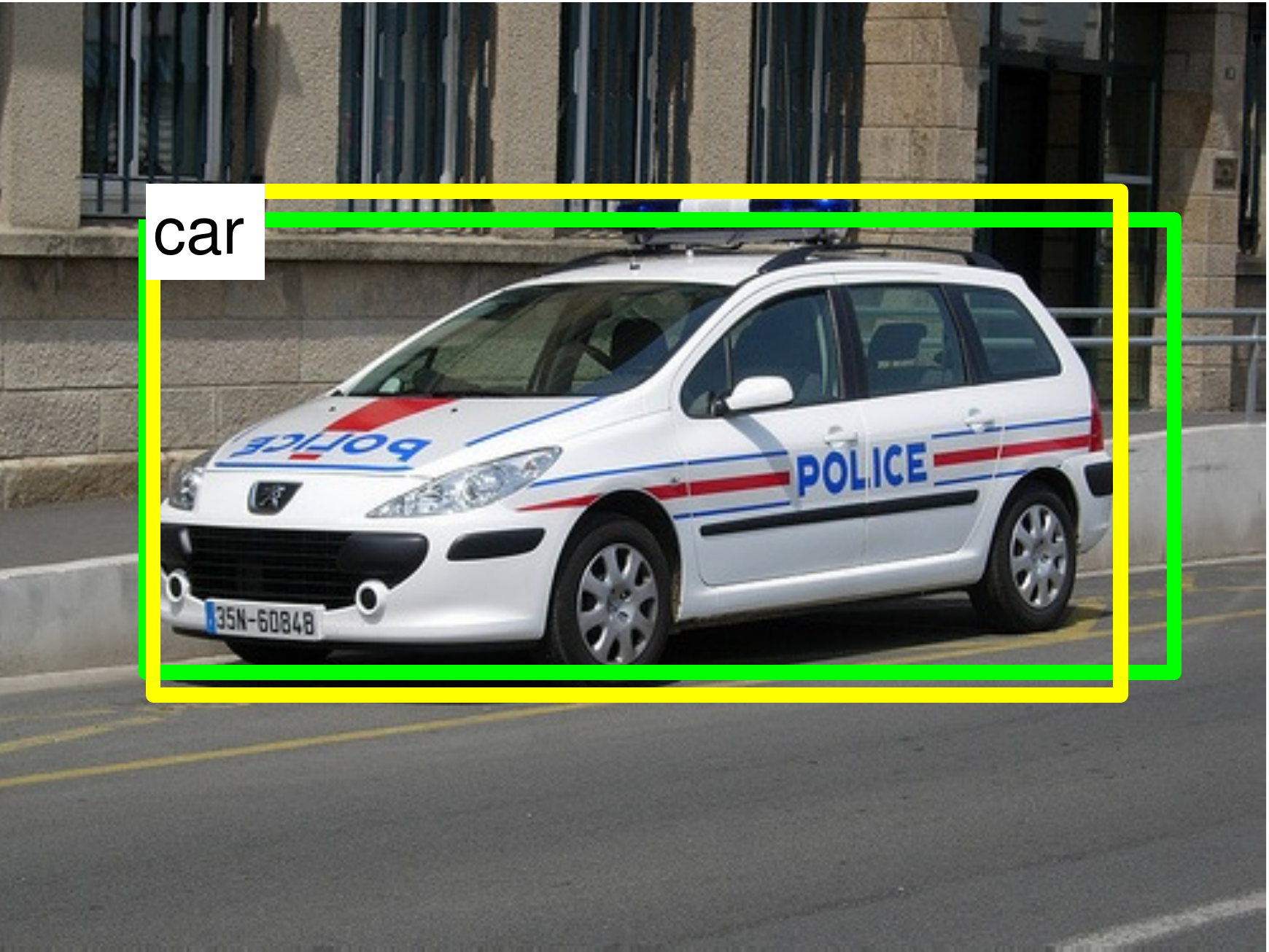}}\hspace{.3em}%
\subfloat{\includegraphics[width=0.18\textwidth,height=0.10\textwidth]{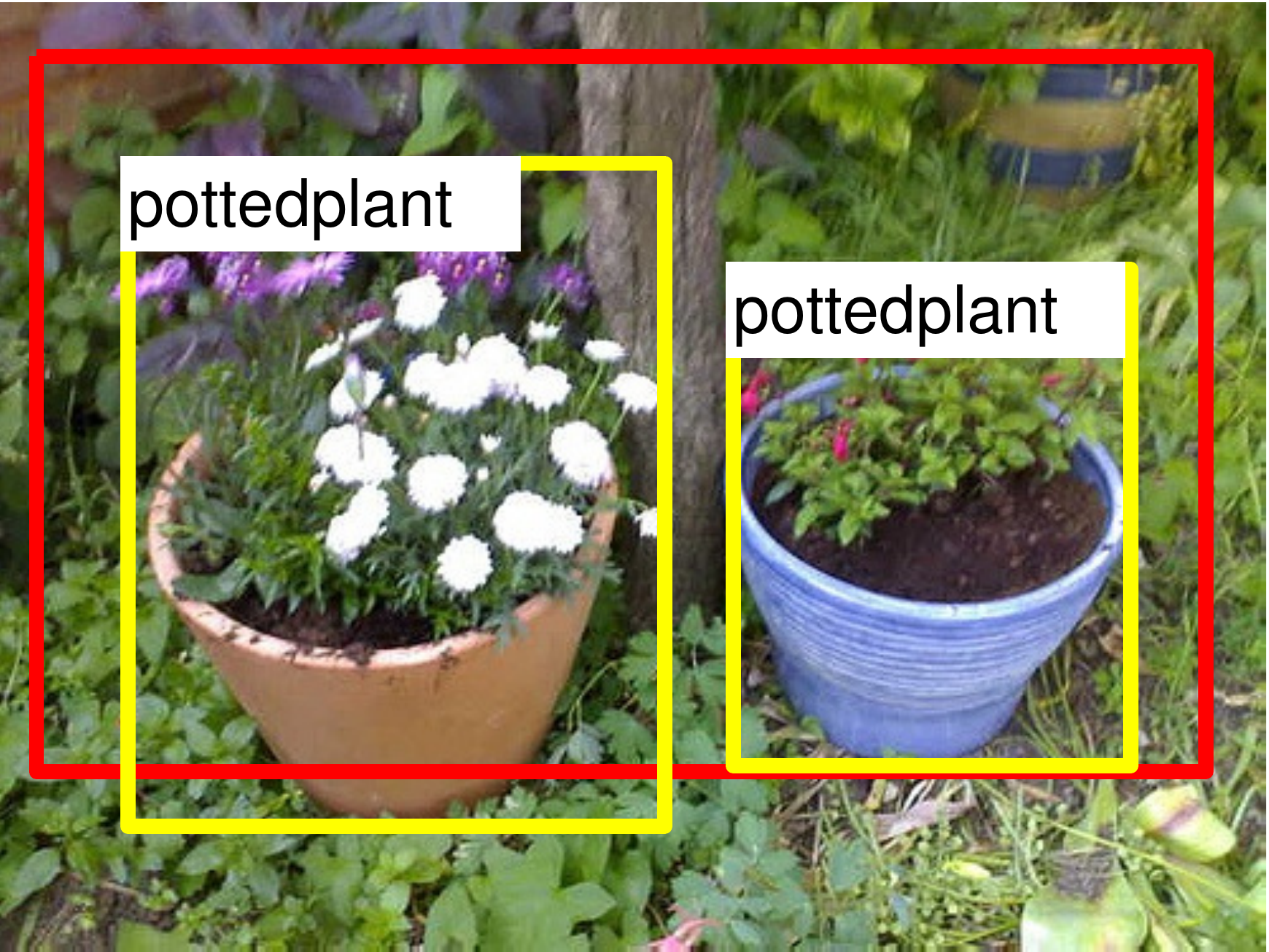}}\hspace{.3em}%
\subfloat{\includegraphics[width=0.18\textwidth,height=0.10\textwidth]{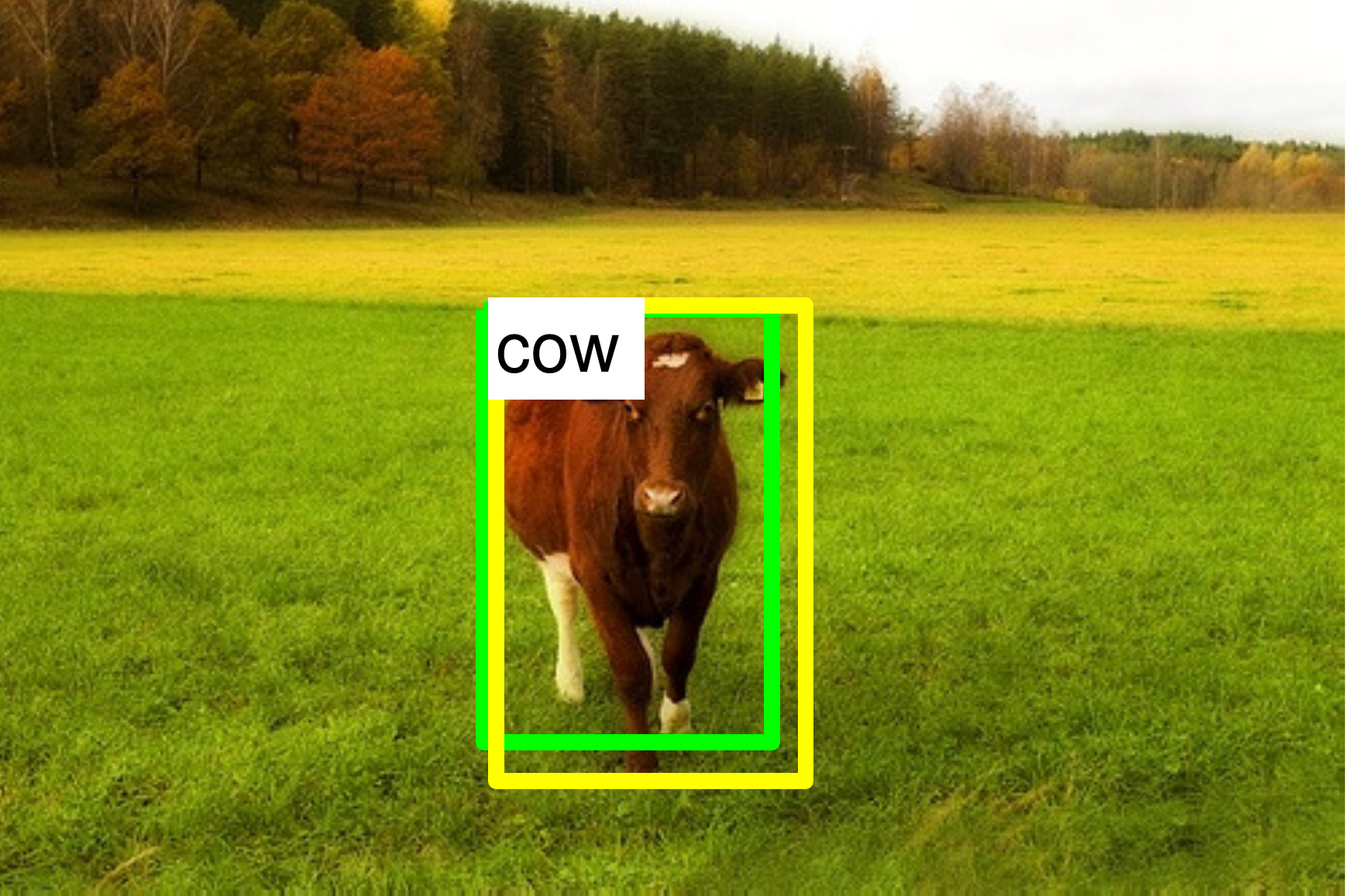}}\hspace{.3em}%
\subfloat{\includegraphics[width=0.18\textwidth,height=0.10\textwidth]{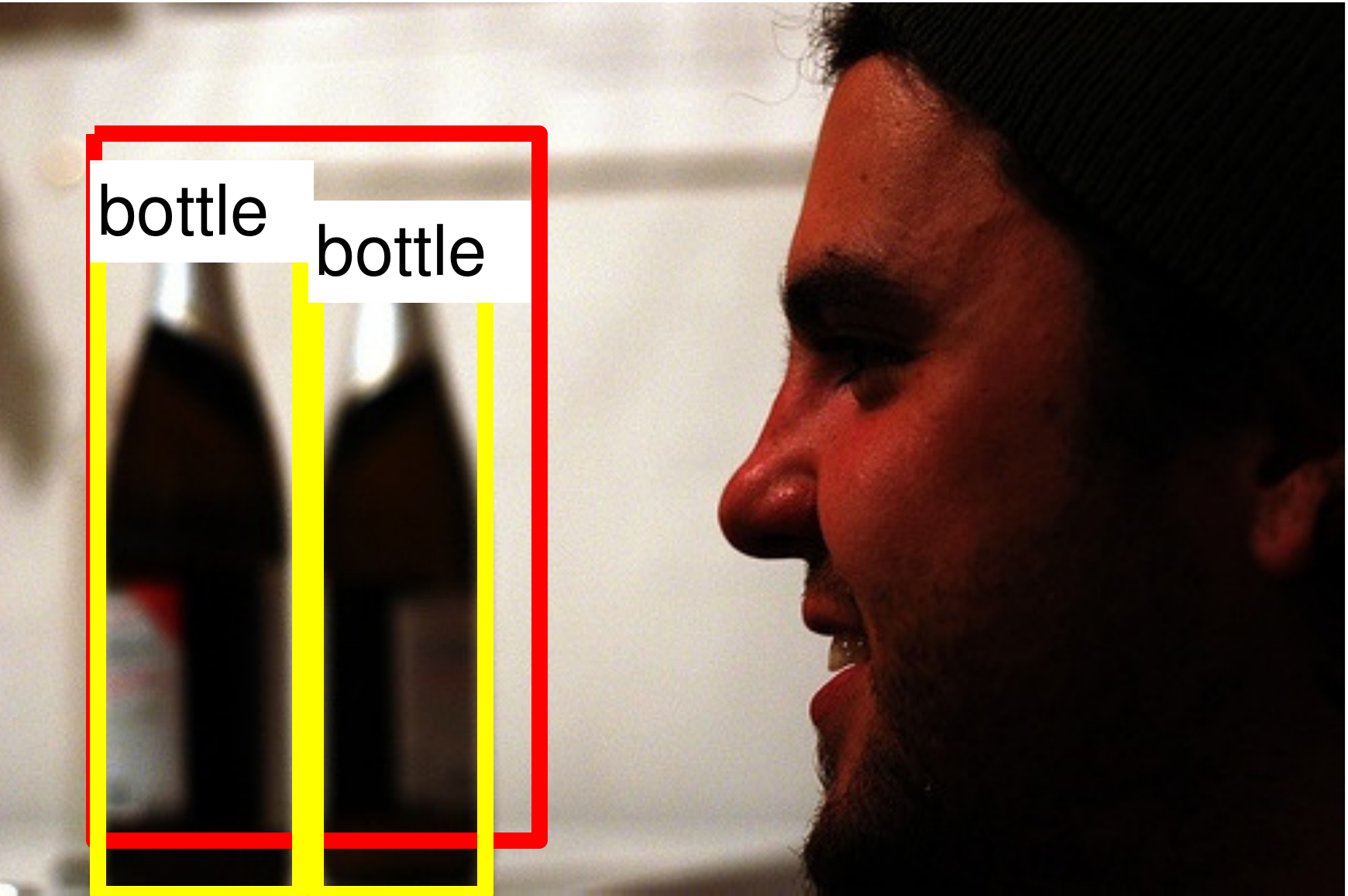}}\vspace{-.75em}\\%
\subfloat{\includegraphics[width=0.18\textwidth,height=0.10\textwidth]{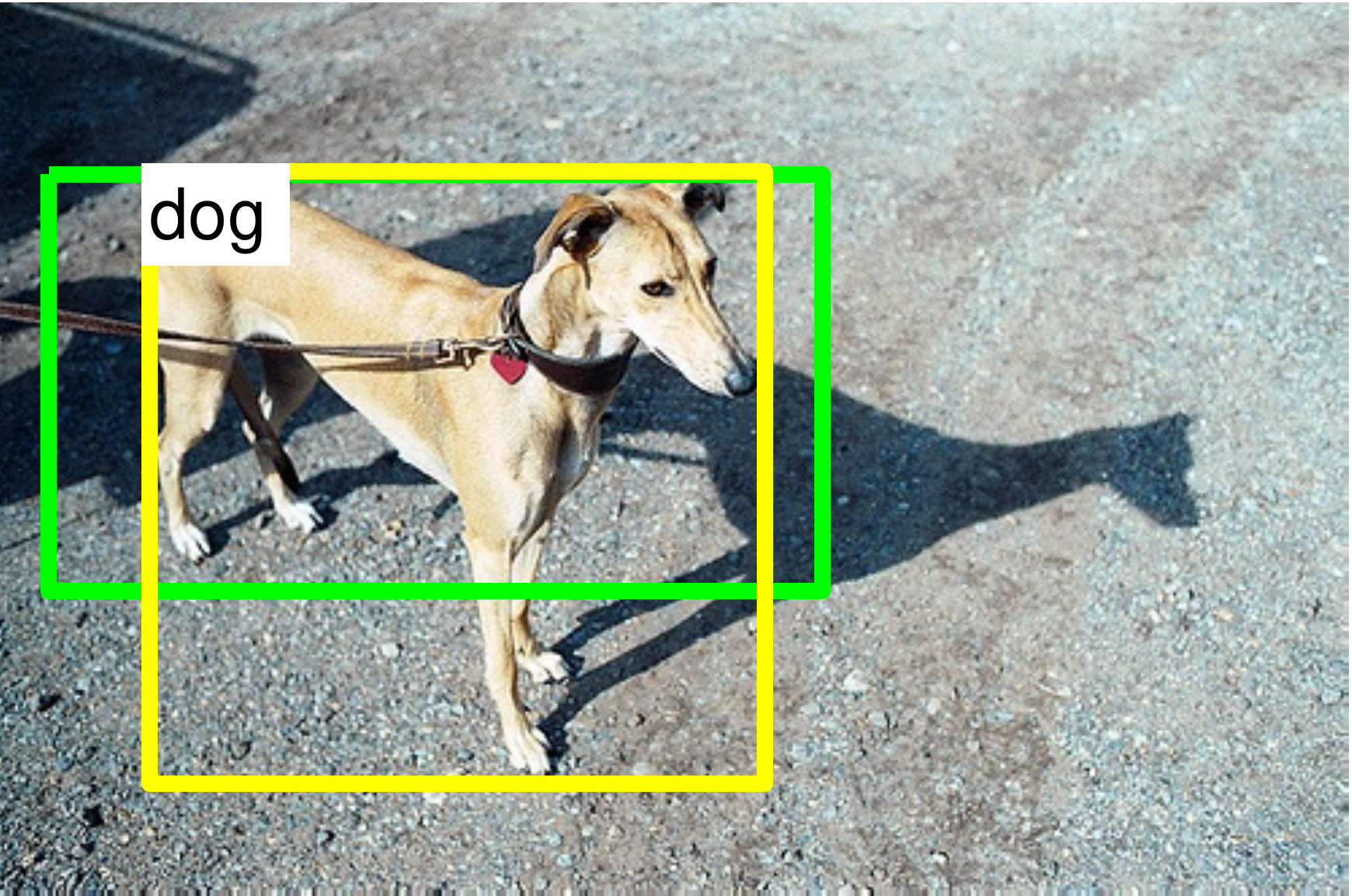}}\hspace{.3em}%
\subfloat{\includegraphics[width=0.18\textwidth,height=0.10\textwidth]{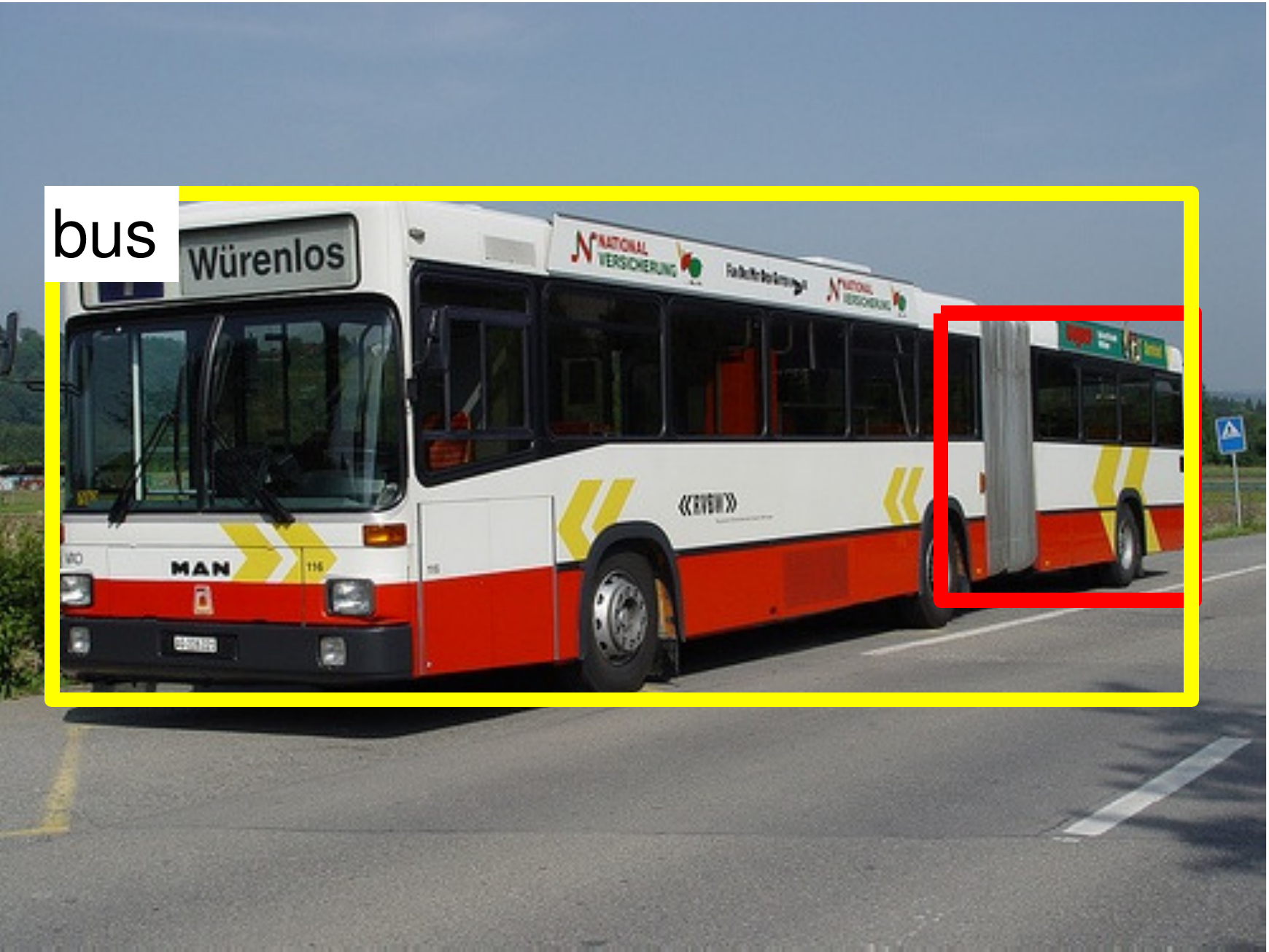}}\hspace{.3em}%
\subfloat{\includegraphics[width=0.18\textwidth,height=0.10\textwidth]{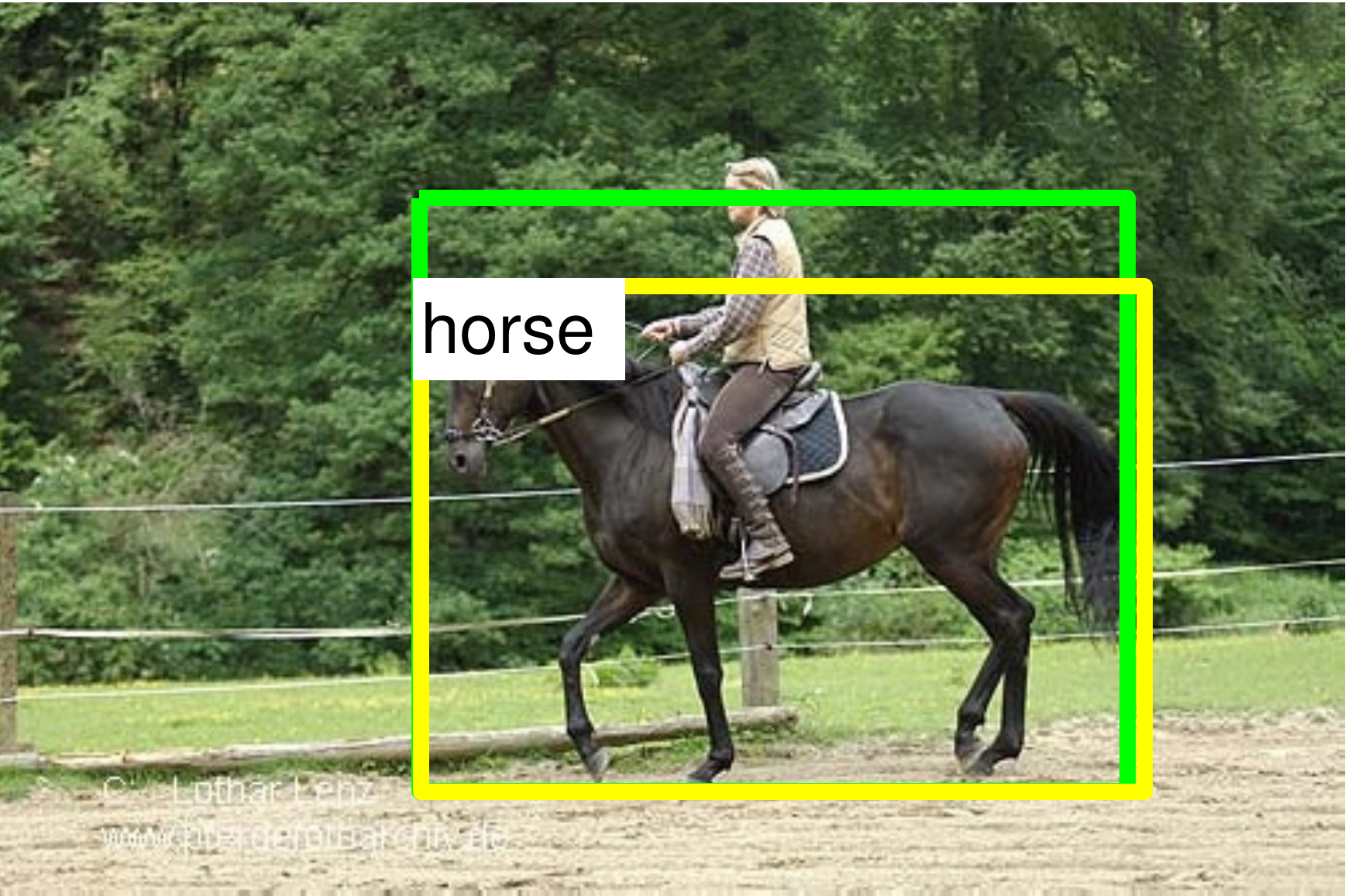}}\hspace{.3em}%
\subfloat{\includegraphics[width=0.18\textwidth,height=0.10\textwidth]{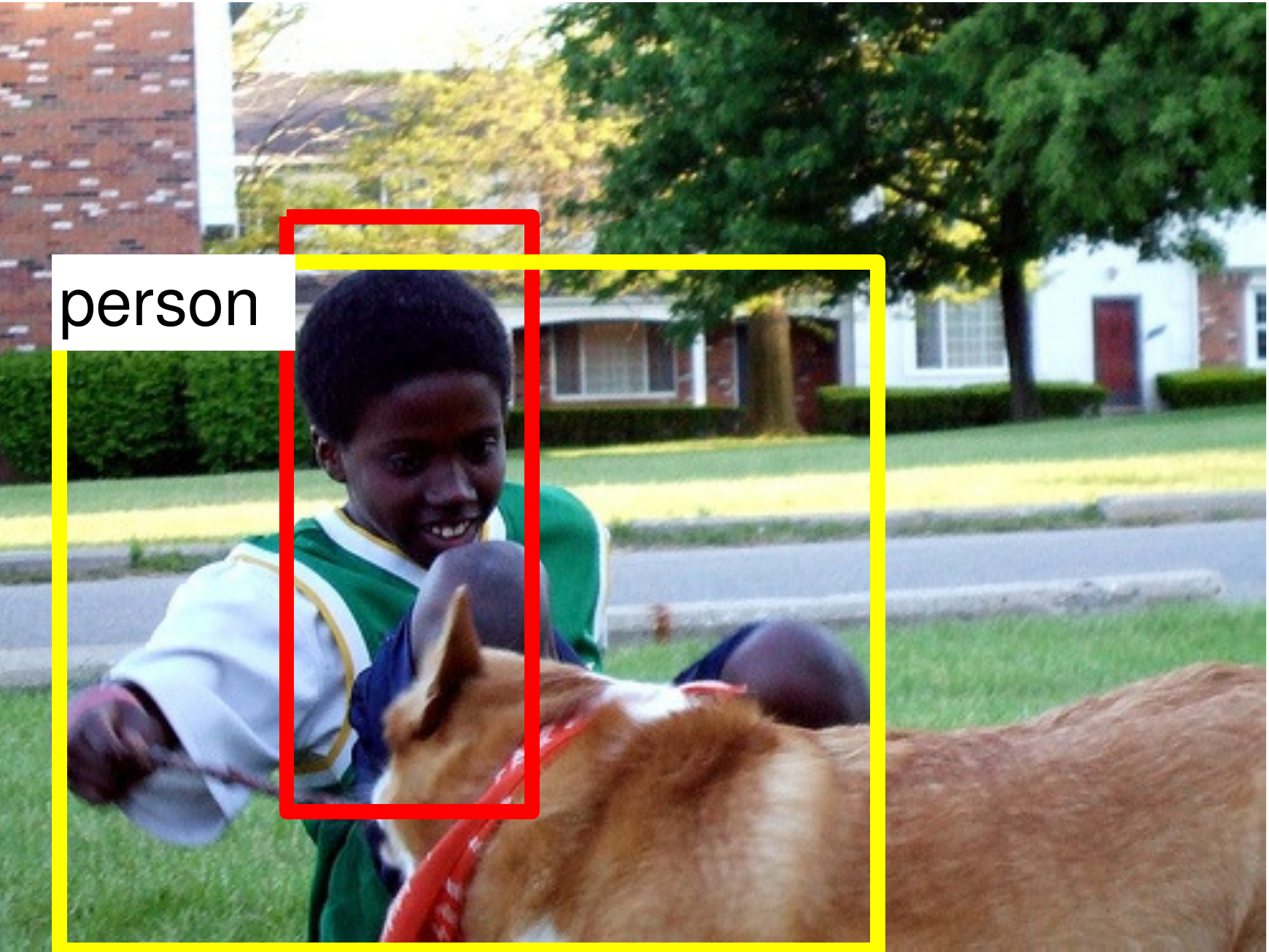}}\hspace{.3em}%
\subfloat{\includegraphics[width=0.18\textwidth,height=0.10\textwidth]{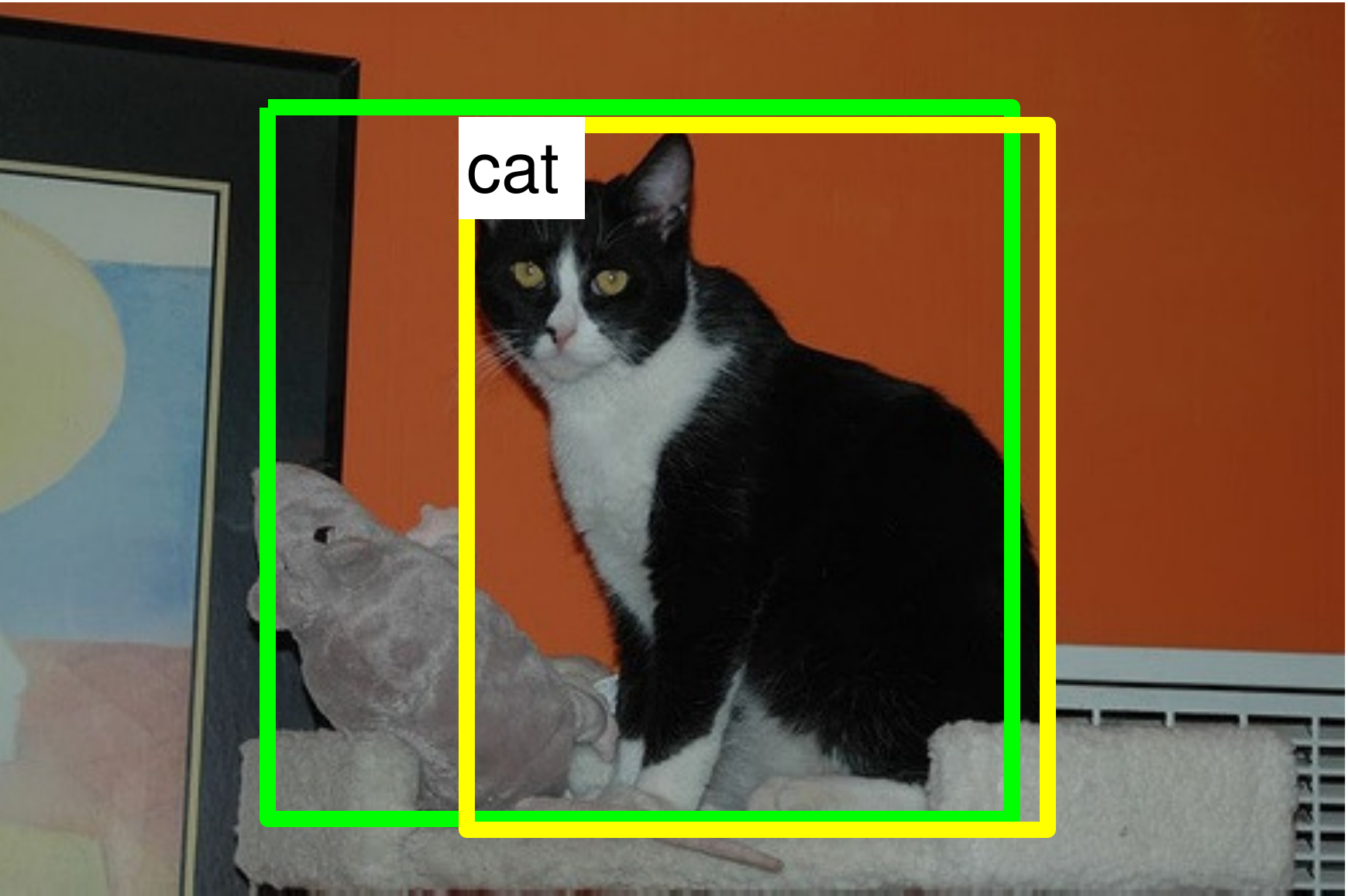}}\vspace{-.75em}\\%
\subfloat{\includegraphics[width=0.18\textwidth,height=0.10\textwidth]{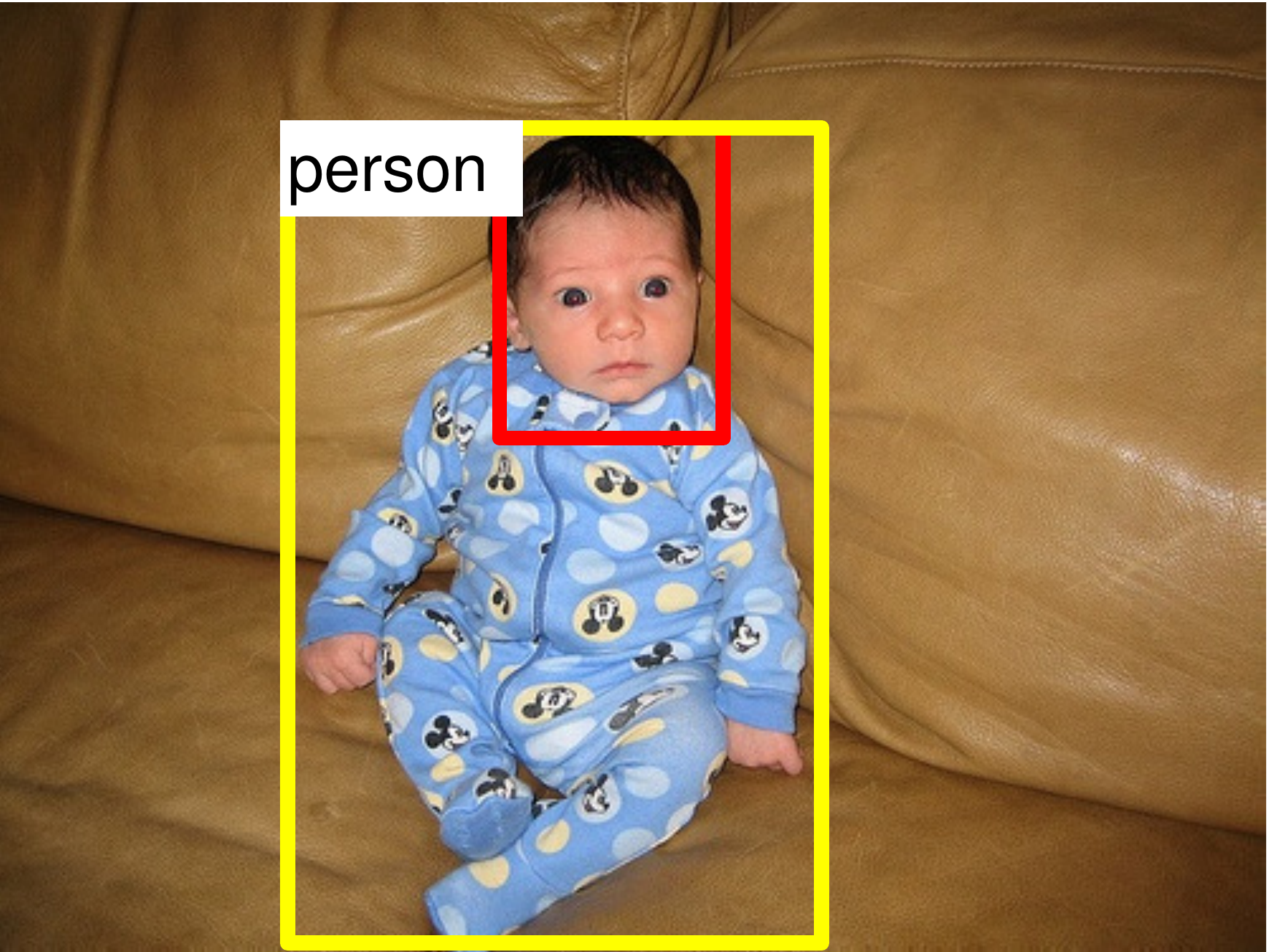}}\hspace{.3em}%
\subfloat{\includegraphics[width=0.18\textwidth,height=0.10\textwidth]{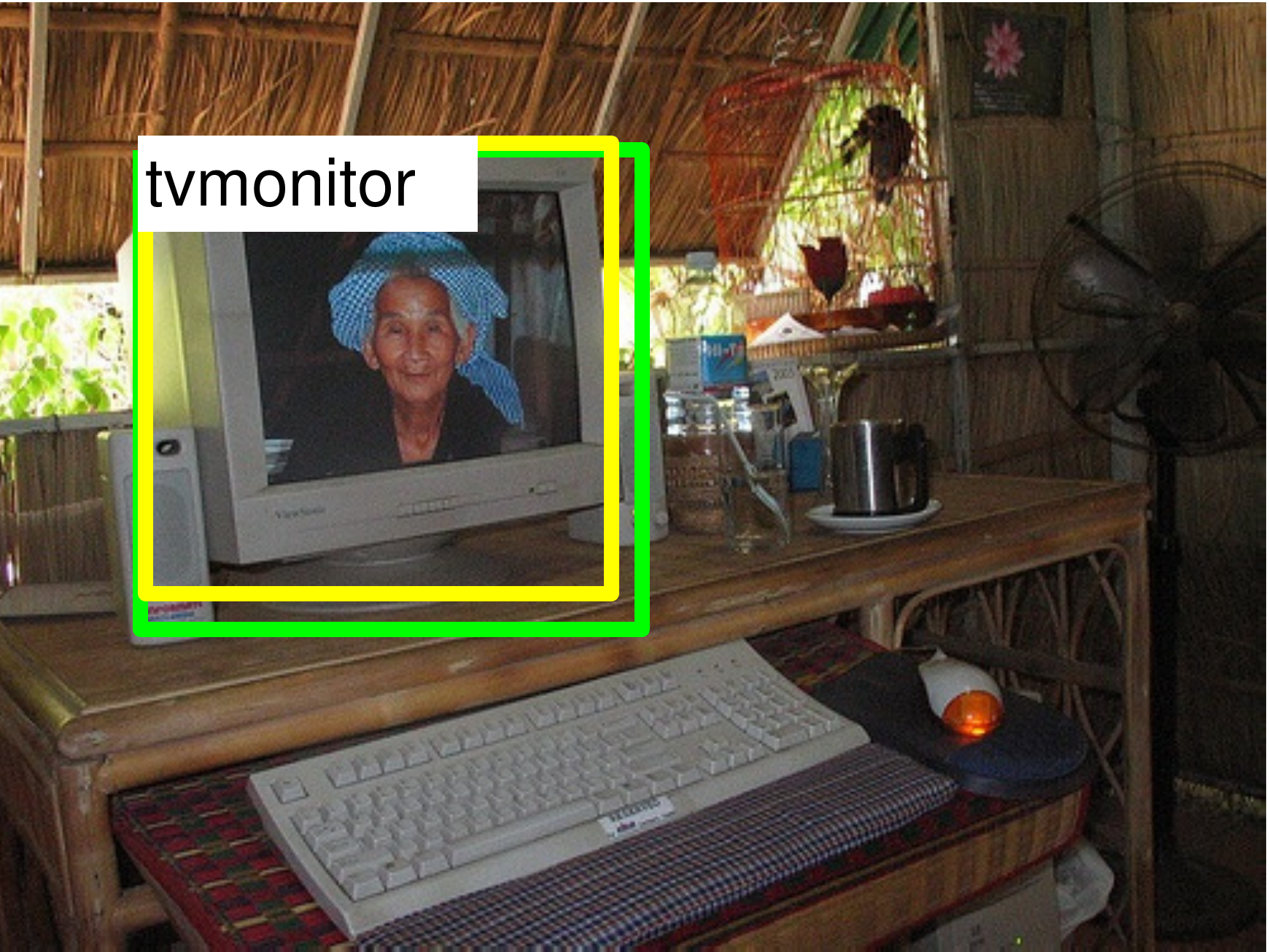}}\hspace{.3em}%
\subfloat{\includegraphics[width=0.18\textwidth,height=0.10\textwidth]{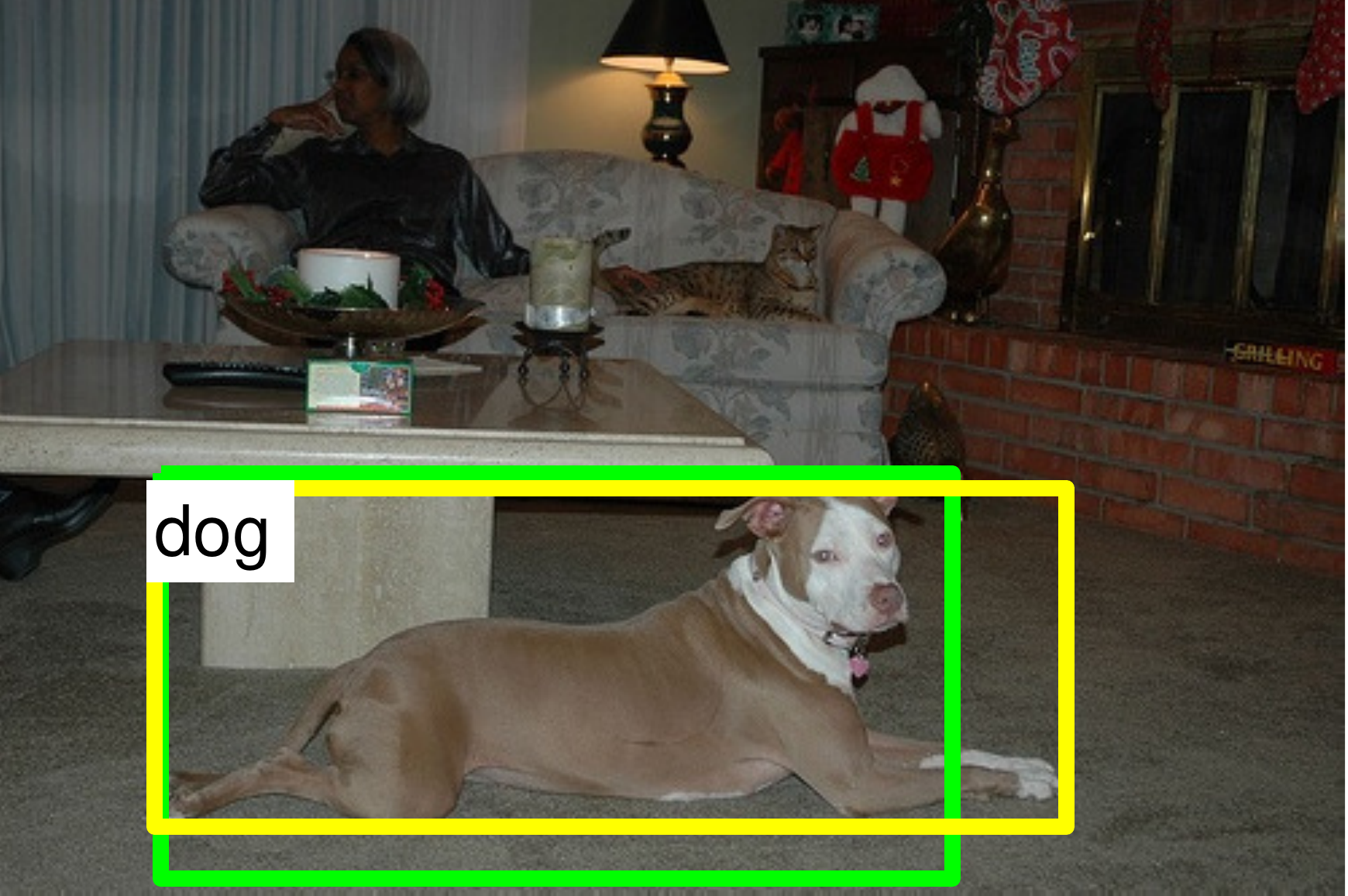}}\hspace{.3em}%
\subfloat{\includegraphics[width=0.18\textwidth,height=0.10\textwidth]{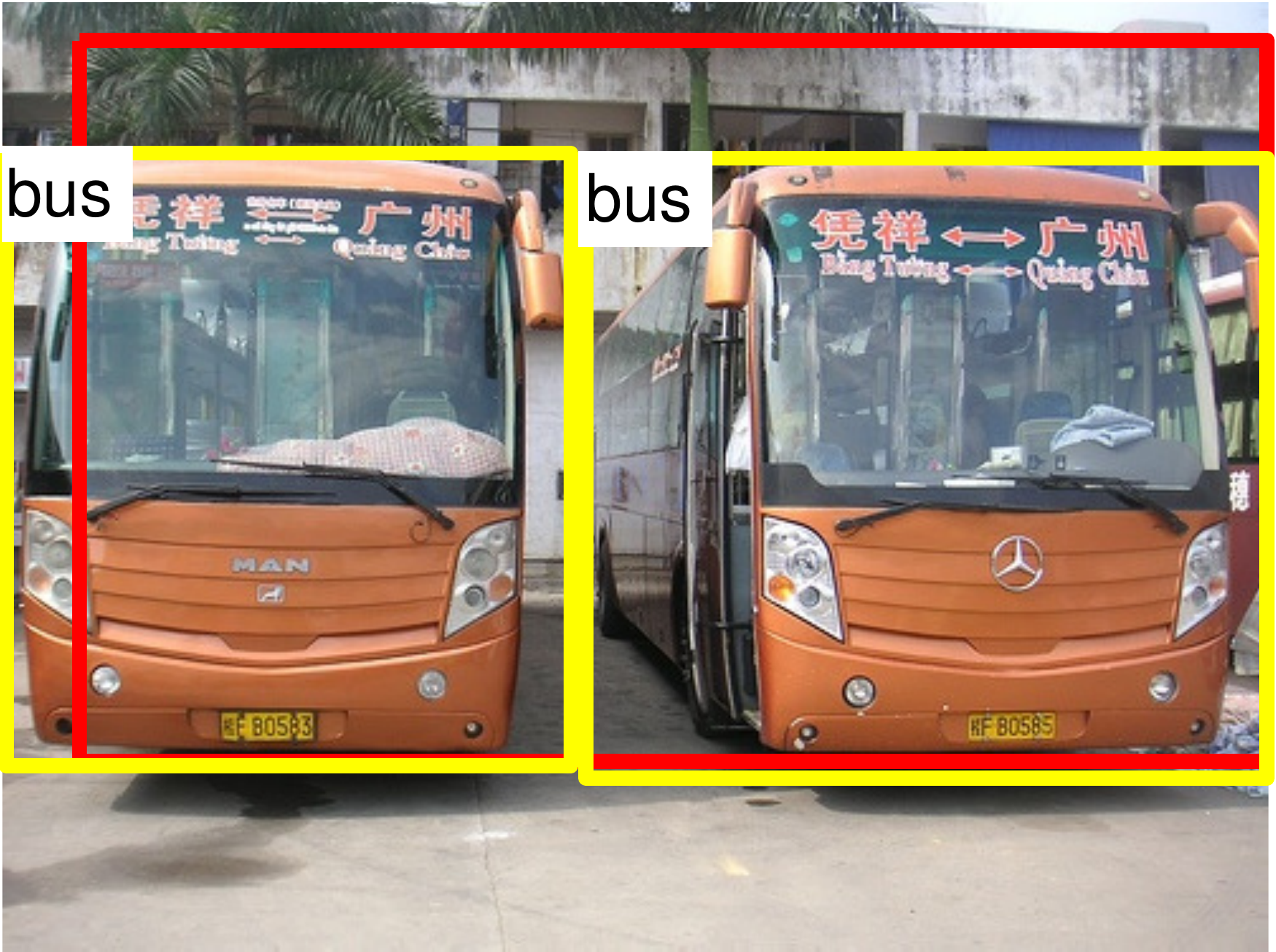}}\hspace{.3em}%
\subfloat{\includegraphics[width=0.18\textwidth,height=0.10\textwidth]{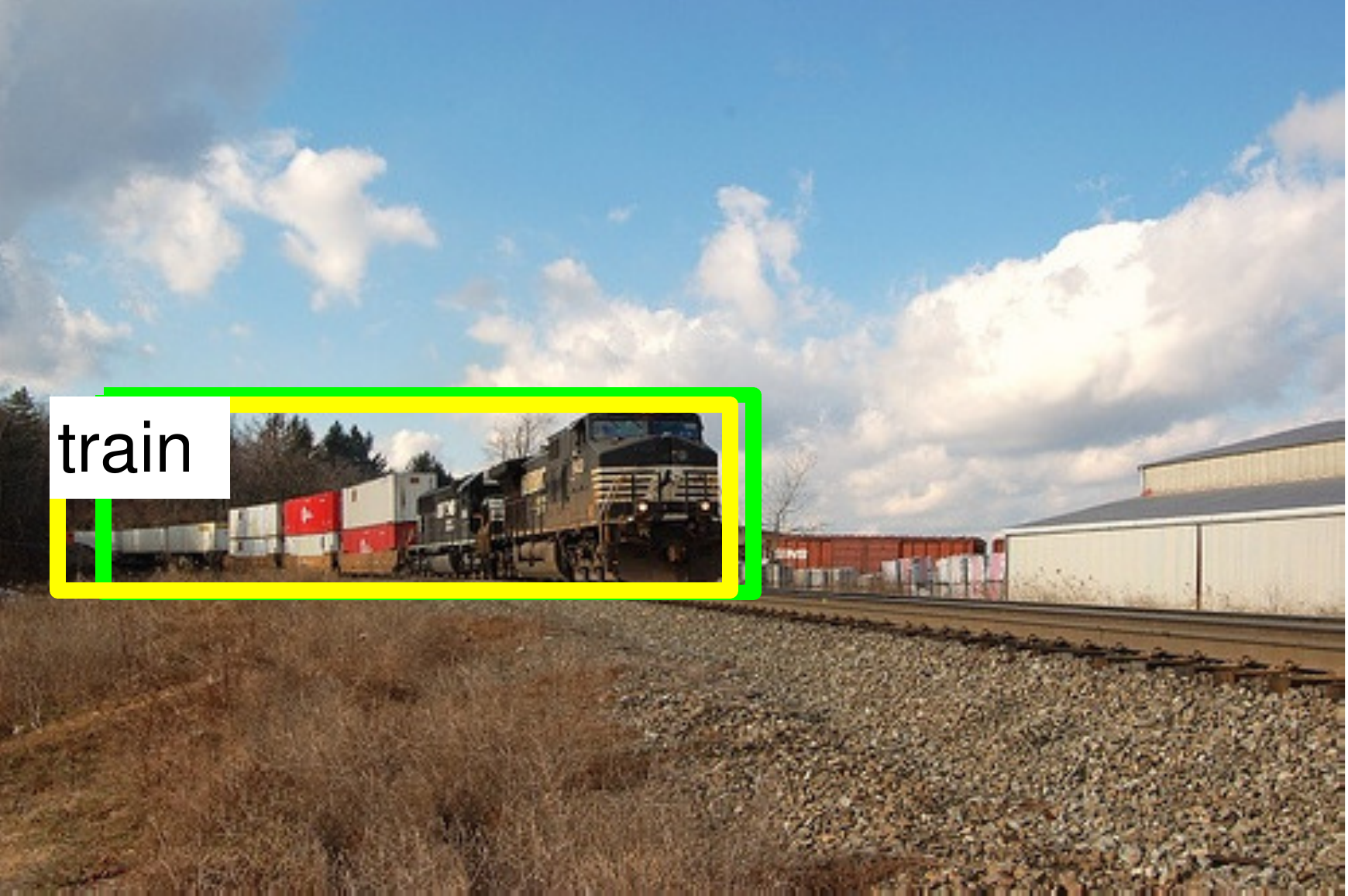}}\vspace{-1.6em}
\end{center}
   \caption{Examples of success (green boxes) and failure (red boxes) cases on the PASCAL VOC 2007 \emph{test} set using K-EM + VGG. Yellow boxes indicates ground-truth annotations. Best viewed in color.}
\label{fig:samples}
\end{figure*}

While our method outperforms the current state-of-the-art approaches in terms of mAP or average CorLoc for all the 20 categories, our performance is not as strong in chair, diningtable, person and pottedplant categories. Sample detection results are illustrated in Fig \ref{fig:samples}. We apply the detector error analysis tool from Hoiem \etal \cite{hoiem2012diagnosing} for these four categories, as shown in Fig \ref{fig:error}. It can be noted that the majority of detection failures comes from failed localization. Like previous methods \cite{Li_2016_CVPR,Bilen_2016_CVPR}, our system often focus on discriminative, less variable object part (\eg, person face) instead of the whole object, due to the high variance of appearance of the whole object. We believe that we can improve our performance by incorporating additional cue about the whole object.

\begin{figure}[t]
\begin{center}
\includegraphics[width=.95\linewidth]{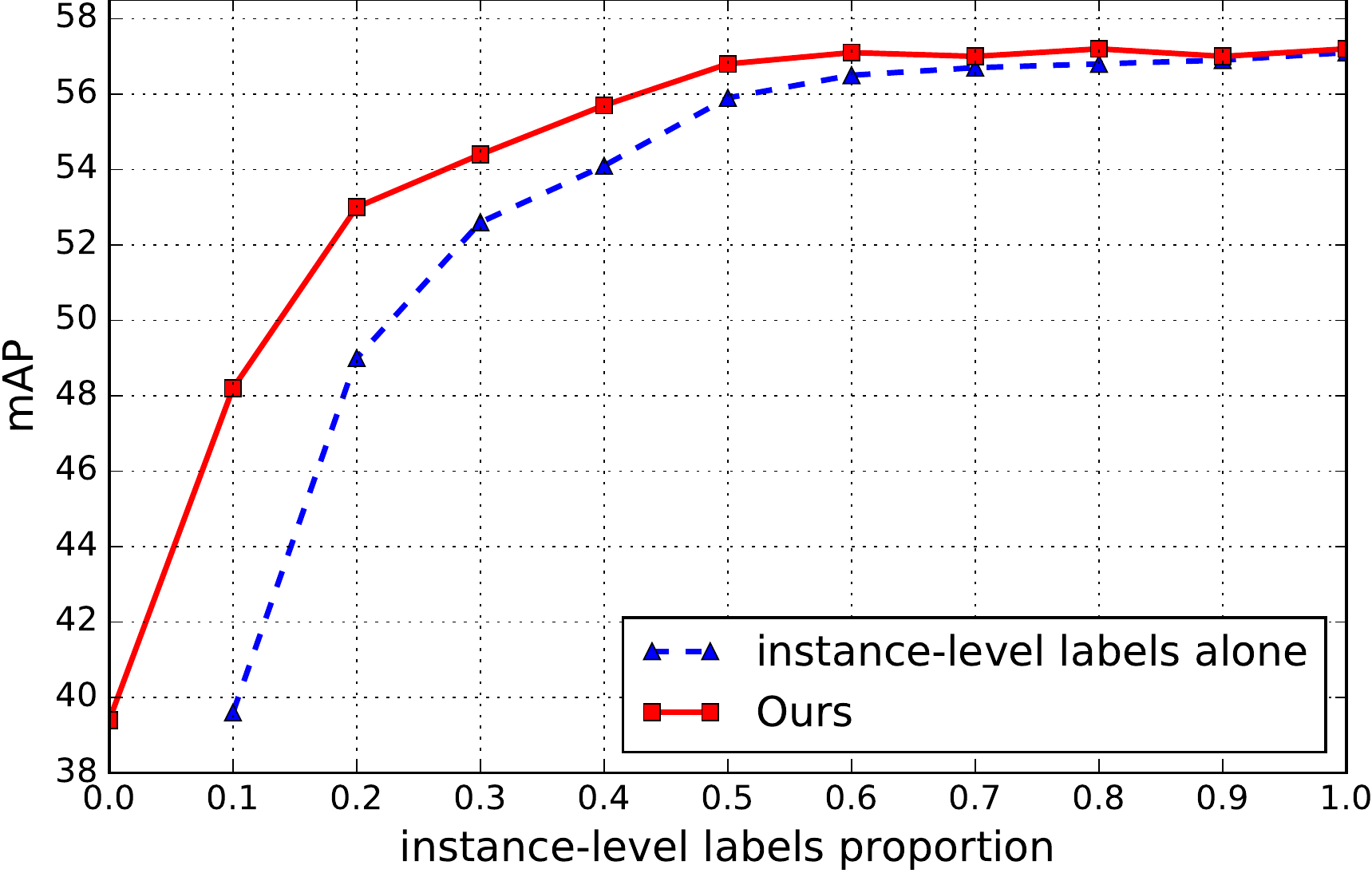}\vspace{-1.2em}
\end{center}
   \caption{Semi supervised detection performance (mAP) on the PASCAL VOC 2007 \emph{test} set.}
\label{fig:semiresult}
\end{figure}

\vspace{.65em}

\noindent{\bf Ablation studies}\quad Our best model, K-EM using VGG, takes WSDDN-ENS-VGG as pre-training. With the EM algorithm, we obtain detection performance improvement on the PASCAL VOC 2007 \emph{test} set from 39.3\% to 45.0\% mAP. For AlexNet, we train a WSDDN initialized from AlexNet, and take this WSDDN as pre-training. Using AlexNet as the base CNN architecture, we improve detection performance from 33.7\% to 39.4\% mAP with the EM algorithm. There are also consistent improvements in terms of CorLoc on the PASCAL VOC 2007 \emph{trainval} set, if we train object detector using EM algorithm after the pre-training step. The result demonstrates the EM algorithm can help the CNN to select better proposals, and learn a better object appearance model. We also compare K-EM and Hard-EM. Using AlexNet, the K-EM setting obtains 1.3\% mAP and 1.1\% CorLoc improvement over the Hard-EM setting. This can be explained by the fact that Hard-EM greedily keeps only one term with largest $P(\bm y|\bm x;\bm\theta')$ in \eqref{eqn:completedatalogweakly}, discarding too much information. In our experiments, setting $K=1000$ only gives marginal performance improvement (about 0.1\% mAP), while significantly increases computational cost. So we report K-EM results under $K=100$. Applying bounding box regression also gives 2.7\% mAP points improvement, demonstrating that recent techniques on fully supervised detection problem can also improve weakly supervised detection performance. It's easy to integrate those techniques into our method, since the core of our M-step is solving a fully supervised detection problem. We also evaluate the effect of different object proposal generators. We compare SelectiveSearch \cite{uijlings2013selective} and Edge Boxes \cite{zitnick2014edge}. In our experiments, by training detector with Edge Boxes, typically about 1.2\% mAP improvement on the PASCAL VOC 2007 \emph{test} set can be obtained over SelectiveSearch. Furthermore, we get very poor performance of 3.4\% mAP if removing the WSDDN pre-training step, validating the importance of WSDDN pre-training.

\subsection{Semi supervised detection results}\label{sec:SSDresults}
We evaluate our method on the PASCAL VOC 2007 benchmark, using image-level labels in combination with instance-level labels for training. For each category in the training set, some of images have instance-level labels, while other images only have image-level labels. In the pre-training step, we use only image-level labels, as in Section \ref{sec:WSDresults}. After that, we use both image-level labels and instance-level labels in the EM algorithm. For simplicity, we use K-EM ($K=100$) with AlexNet in all semi supervised experiments. 

\begin{figure*}[t!]
\begin{center}
\captionsetup[subfigure]{labelformat=empty}
\subfloat{\includegraphics[width=0.18\textwidth,height=0.10\textwidth]{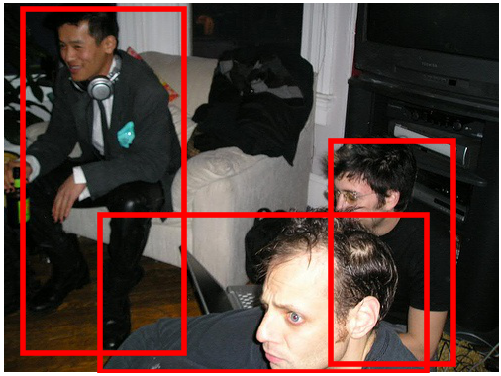}}\hspace{.3em}%
\subfloat{\includegraphics[width=0.18\textwidth,height=0.10\textwidth]{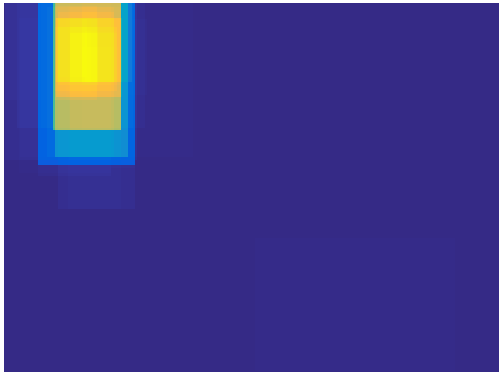}}\hspace{.3em}%
\subfloat{\includegraphics[width=0.18\textwidth,height=0.10\textwidth]{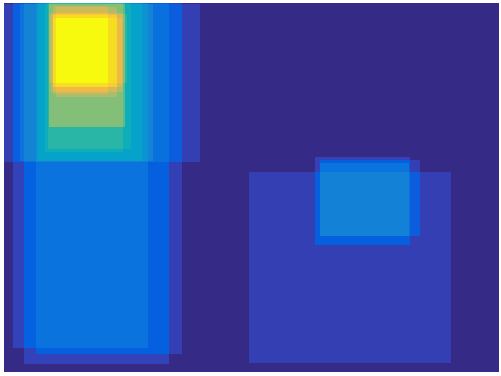}}\hspace{.3em}%
\subfloat{\includegraphics[width=0.18\textwidth,height=0.10\textwidth]{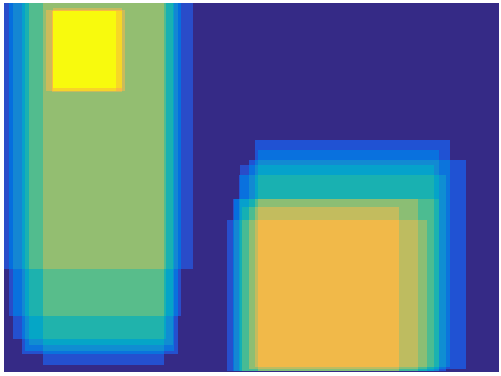}}\hspace{.3em}%
\subfloat{\includegraphics[width=0.18\textwidth,height=0.10\textwidth]{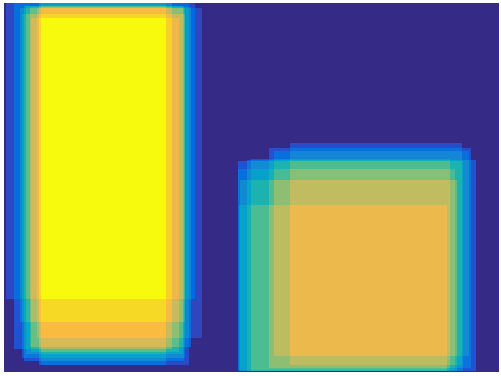}}\vspace{-.75em}\\%
\subfloat{\includegraphics[width=0.18\textwidth,height=0.10\textwidth]{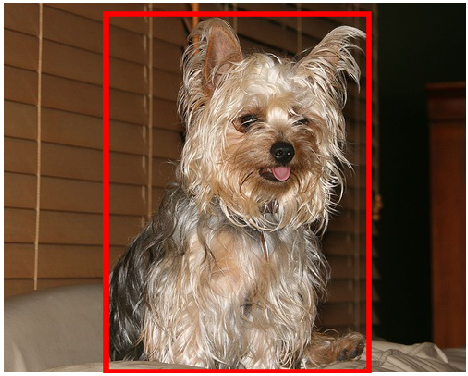}}\hspace{.3em}%
\subfloat{\includegraphics[width=0.18\textwidth,height=0.10\textwidth]{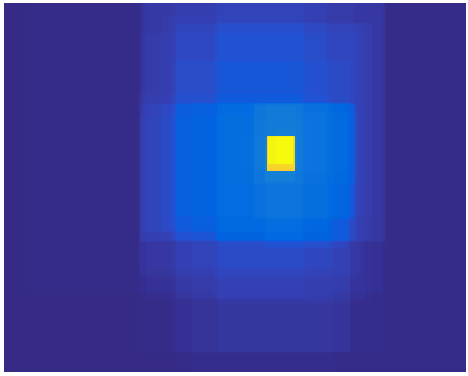}}\hspace{.3em}%
\subfloat{\includegraphics[width=0.18\textwidth,height=0.10\textwidth]{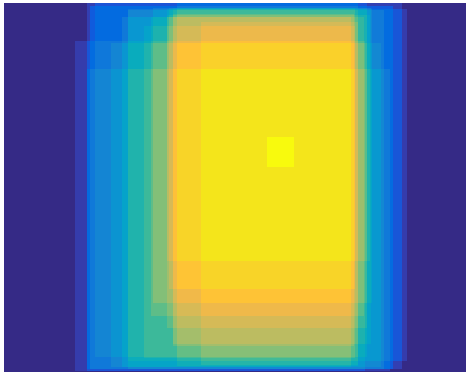}}\hspace{.3em}%
\subfloat{\includegraphics[width=0.18\textwidth,height=0.10\textwidth]{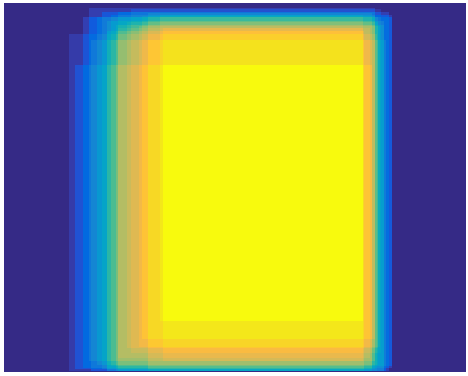}}\hspace{.3em}%
\subfloat{\includegraphics[width=0.18\textwidth,height=0.10\textwidth]{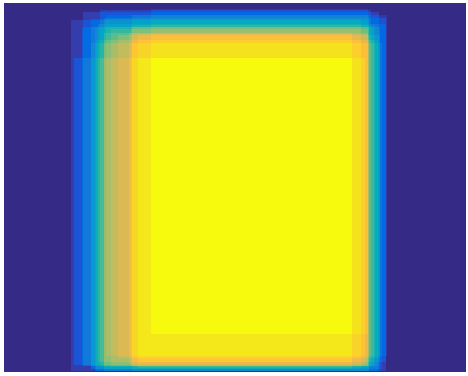}}\vspace{-.75em}\\%
\subfloat{\includegraphics[width=0.18\textwidth,height=0.10\textwidth]{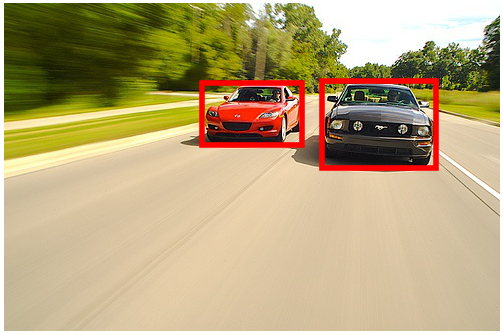}}\hspace{.3em}%
\subfloat{\includegraphics[width=0.18\textwidth,height=0.10\textwidth]{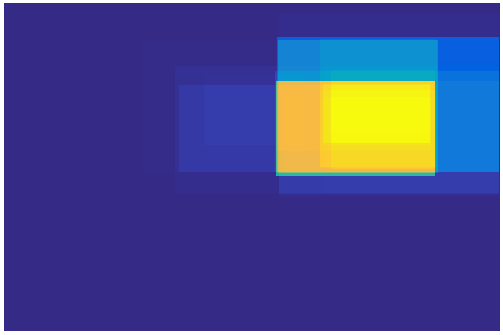}}\hspace{.3em}%
\subfloat{\includegraphics[width=0.18\textwidth,height=0.10\textwidth]{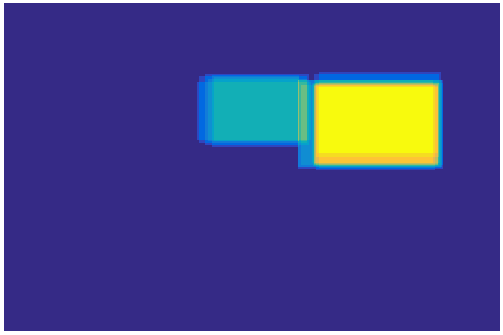}}\hspace{.3em}%
\subfloat{\includegraphics[width=0.18\textwidth,height=0.10\textwidth]{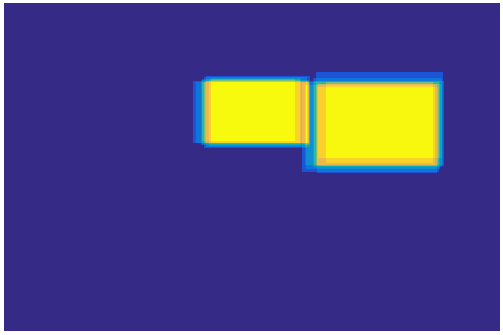}}\hspace{.3em}%
\subfloat{\includegraphics[width=0.18\textwidth,height=0.10\textwidth]{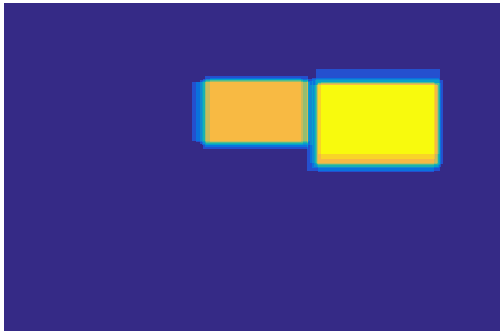}}\vspace{-.75em}\\%
\subfloat{\includegraphics[width=0.18\textwidth,height=0.10\textwidth]{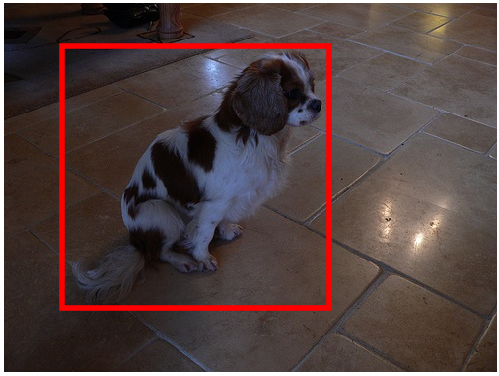}}\hspace{.3em}%
\subfloat[pre-training]{\includegraphics[width=0.18\textwidth,height=0.10\textwidth]{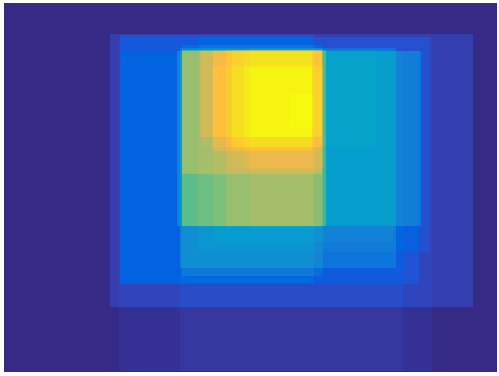}}\hspace{.3em}%
\subfloat[1\textsuperscript{st} M-step]{\includegraphics[width=0.18\textwidth,height=0.10\textwidth]{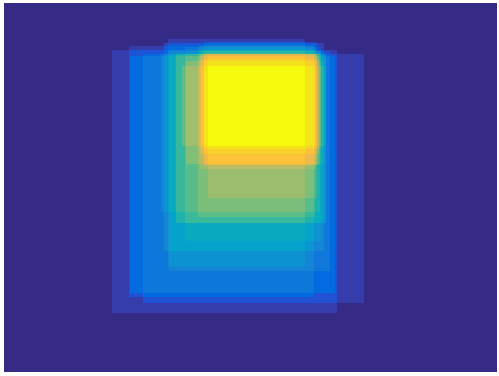}}\hspace{.3em}%
\subfloat[2\textsuperscript{nd} M-step]{\includegraphics[width=0.18\textwidth,height=0.10\textwidth]{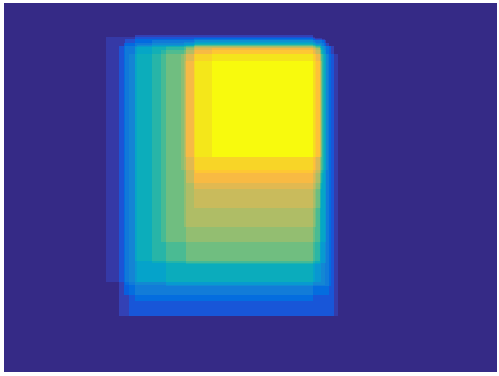}}\hspace{.3em}%
\subfloat[3\textsuperscript{rd} M-step]{\includegraphics[width=0.18\textwidth,height=0.10\textwidth]{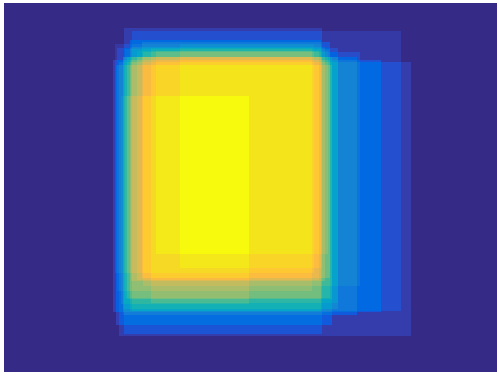}}\vspace{-1.6em}%
\end{center}
   \caption{Examples of detection response maps on weakly annotated training images using K-EM + VGG. The first column shows the ground truth. The second column shows the weighted sum of window scores from WSDDN pre-training. The third to fifth columns show the weighted sum of window scores from our method, at different iterations. Best viewed in color.}
\label{fig:emweight}
\end{figure*}

Results are summarized in Fig \ref{fig:semiresult}. By training with 40\% instance-level labels and 60\% image-level labels, we achieve 55.7\% mAP, which is only 1.4\% inferior the fully supervised Fast RCNN (100\% instance-level labels, 57.1\% mAP). Note the performance is always lower if we train detector from image-level labels alone, as shown in Fig~\ref{fig:semiresult}. Fig~\ref{fig:emweight} shows the response maps on weakly annotated training images (with 50\% instance-level labels available). Our method progressively refine the localization during the training.

\section{Related Work}\label{sec:related}

\noindent{\bf Weakly Supervised Detection.}\quad Many existing methods \cite{song2014learning,cinbis2014multi,Li_2016_CVPR,song2014weakly,song2014learning,siva2012defence,bilen2014weakly} formulate WSD as a \emph{Multiple Instance Learning}~(MIL) problem. They usually use MI-SVM to mine positive object proposals, and develop better initialization and optimization strategies in order to prevent poor local extrema. Li \etal \cite{Li_2016_CVPR} propose a two-step domain adaptation approach. They transfer the classifier from the ImageNet 1000 categories to the PASCAL VOC 20 categories, and filter object proposals class-by-class. Then they apply MI-SVM on a cleaner collection of class-specific object proposals. Finally they use mined confident object candidates to train a Fast RCNN. Despite the fact that we address WSD in a different point of view from \cite{yang2016weakly,Li_2016_CVPR} (they mine positive patches while we estimate probability of objects over all possible locations), their methods actually end up to be special cases of our method. From our point of view, \cite{yang2016weakly,Li_2016_CVPR} apply Hard-EM approximation after carefully initialization, and perform only one EM iteration (\ie, a single E-step followed by a single M-step). Hard-EM discards too much information in the E-step, and the CNN weights are usually not fully converged in single EM iteration. Bilen~\etal \cite{bilen2015weakly} also assign instance-level labels in a smooth way. They help the optimization by enforcing a \emph{soft} similarity between each possible location in the image and a reduced set of \emph{exemplars}. WSDDN \cite{Bilen_2016_CVPR} and subsequent work \cite{kantorov2016contextlocnet,yang2016weakly} achieve state-of-the-art performance, using an end-to-end, two-stream CNN architecture to perform region selection and appearance model learning simultaneously. The main differences between our methods and existing WSD methods are: (1) Our method can be applied to semi supervised setting. (2) From our point of view, these methods can either be a pre-training step or a special case of our framework. We believe it's possible to gain potential performance improvement by integrating these methods into our framework.

\vspace{.65em}

\noindent{\bf Semi Supervised Detection.}\quad There are several previous works \cite{hoffman2014lsda,hoffman2015detector,Tang_2016_CVPR} which focus on combing image-level labels and instance-level labels in object detection. They assume the existence of strongly annotated categories, and transfer knowledge from these strongly annotated categories to weakly annotated categories. For example, if we want to detect the 20 PASCAL VOC categories \cite{pascal-voc-2007}, some additional strongly annotated categories, which are not in the 20 PASCAL VOC categories, are required. Our work differs these methods in that, we focus on a different setting, where both image-level labels and instance-level labels are in the same category, thus no additional strongly annotated categories are required.

\section{Conclusion}\label{sec:conclusion}
We present an EM based method to train object detector from image-level labels using deep convolutional neural networks. We treat instance-level labels as missing values. Our method can learn detector from either image-level labels alone or in combination with some instance-level labels. Using image-level labels solely, our method achieves 46.1\% mAP on the PASCAL VOC 2007 test set, outperforming the current state-of-the-art weakly supervised detection approaches. Having access to little instance-level labels, our method can almost match the performance of the fully supervised Fast RCNN. Our results show that by exploiting weakly annotated images, excellent detection performance can be obtained with less annotation effort.


\newpage
{\small
\bibliographystyle{ieee}
\bibliography{ref}

\begin{thebibliography}{10}\itemsep=-1pt

\bibitem{bilen2014weakly}
H.~Bilen, M.~Pedersoli, and T.~Tuytelaars.
\newblock Weakly supervised object detection with posterior regularization.
\newblock In {\em BMVC}, volume~3, 2014.

\bibitem{bilen2015weakly}
H.~Bilen, M.~Pedersoli, and T.~Tuytelaars.
\newblock Weakly supervised object detection with convex clustering.
\newblock In {\em CVPR}, pages 1081--1089, 2015.

\bibitem{Bilen_2016_CVPR}
H.~Bilen and A.~Vedaldi.
\newblock Weakly supervised deep detection networks.
\newblock In {\em CVPR}, June 2016.

\bibitem{cinbis2014multi}
R.~G. Cinbis, J.~Verbeek, and C.~Schmid.
\newblock Multi-fold mil training for weakly supervised object localization.
\newblock In {\em CVPR}, pages 2409--2416. IEEE, 2014.

\bibitem{deselaers2012weakly}
T.~Deselaers, B.~Alexe, and V.~Ferrari.
\newblock Weakly supervised localization and learning with generic knowledge.
\newblock {\em IJCV}, 100(3):275--293, 2012.

\bibitem{diba2016weakly}
A.~Diba, V.~Sharma, A.~Pazandeh, H.~Pirsiavash, and L.~Van~Gool.
\newblock Weakly supervised cascaded convolutional networks.
\newblock {\em arXiv preprint arXiv:1611.08258}, 2016.

\bibitem{pascal-voc-2007}
M.~Everingham, L.~Van~Gool, C.~K.~I. Williams, J.~Winn, and A.~Zisserman.
\newblock The {PASCAL} {V}isual {O}bject {C}lasses {C}hallenge 2007 {(VOC2007)}
  {R}esults.

\bibitem{girshick2015fast}
R.~Girshick.
\newblock Fast r-cnn.
\newblock In {\em ICCV}, pages 1440--1448, 2015.

\bibitem{girshick2014rich}
R.~Girshick, J.~Donahue, T.~Darrell, and J.~Malik.
\newblock Rich feature hierarchies for accurate object detection and semantic
  segmentation.
\newblock In {\em CVPR}, pages 580--587, 2014.

\bibitem{guillaumin2009tagprop}
M.~Guillaumin, T.~Mensink, J.~Verbeek, and C.~Schmid.
\newblock Tagprop: Discriminative metric learning in nearest neighbor models
  for image auto-annotation.
\newblock In {\em ICCV}, pages 309--316. IEEE, 2009.

\bibitem{he2014spatial}
K.~He, X.~Zhang, S.~Ren, and J.~Sun.
\newblock Spatial pyramid pooling in deep convolutional networks for visual
  recognition.
\newblock In {\em ECCV}, pages 346--361. Springer, 2014.

\bibitem{He_2016_CVPR}
K.~He, X.~Zhang, S.~Ren, and J.~Sun.
\newblock Deep residual learning for image recognition.
\newblock In {\em CVPR}, June 2016.

\bibitem{hoffman2014lsda}
J.~Hoffman, S.~Guadarrama, E.~S. Tzeng, R.~Hu, J.~Donahue, R.~Girshick,
  T.~Darrell, and K.~Saenko.
\newblock Lsda: Large scale detection through adaptation.
\newblock In {\em NIPS}, pages 3536--3544, 2014.

\bibitem{hoffman2015detector}
J.~Hoffman, D.~Pathak, T.~Darrell, and K.~Saenko.
\newblock Detector discovery in the wild: Joint multiple instance and
  representation learning.
\newblock In {\em CVPR}, pages 2883--2891, 2015.

\bibitem{hoiem2012diagnosing}
D.~Hoiem, Y.~Chodpathumwan, and Q.~Dai.
\newblock Diagnosing error in object detectors.
\newblock In {\em ECCV}, pages 340--353. Springer, 2012.

\bibitem{jia2014caffe}
Y.~Jia, E.~Shelhamer, J.~Donahue, S.~Karayev, J.~Long, R.~Girshick,
  S.~Guadarrama, and T.~Darrell.
\newblock Caffe: Convolutional architecture for fast feature embedding.
\newblock {\em arXiv preprint arXiv:1408.5093}, 2014.

\bibitem{kantorov2016contextlocnet}
V.~Kantorov, M.~Oquab, M.~Cho, and I.~Laptev.
\newblock Contextlocnet: Context-aware deep network models for weakly
  supervised localization.
\newblock In {\em ECCV}, pages 350--365. Springer, 2016.

\bibitem{krizhevsky2012imagenet}
A.~Krizhevsky, I.~Sutskever, and G.~E. Hinton.
\newblock Imagenet classification with deep convolutional neural networks.
\newblock In {\em NIPS}, pages 1097--1105, 2012.

\bibitem{lecun1998gradient}
Y.~LeCun, L.~Bottou, Y.~Bengio, and P.~Haffner.
\newblock Gradient-based learning applied to document recognition.
\newblock {\em Proceedings of the IEEE}, 86(11):2278--2324, 1998.

\bibitem{Li_2016_CVPR}
D.~Li, J.-B. Huang, Y.~Li, S.~Wang, and M.-H. Yang.
\newblock Weakly supervised object localization with progressive domain
  adaptation.
\newblock In {\em CVPR}, June 2016.

\bibitem{liu2016ssd}
W.~Liu, D.~Anguelov, D.~Erhan, C.~Szegedy, S.~Reed, C.-Y. Fu, and A.~C. Berg.
\newblock {SSD}: Single shot multibox detector.
\newblock In {\em ECCV}, 2016.

\bibitem{perronnin2010improving}
F.~Perronnin, J.~S{\'a}nchez, and T.~Mensink.
\newblock Improving the fisher kernel for large-scale image classification.
\newblock In {\em ECCV}, pages 143--156. Springer, 2010.

\bibitem{Redmon_2016_CVPR}
J.~Redmon, S.~Divvala, R.~Girshick, and A.~Farhadi.
\newblock You only look once: Unified, real-time object detection.
\newblock In {\em CVPR}, June 2016.

\bibitem{ren2015faster}
S.~Ren, K.~He, R.~Girshick, and J.~Sun.
\newblock Faster r-cnn: Towards real-time object detection with region proposal
  networks.
\newblock In {\em NIPS}, pages 91--99, 2015.

\bibitem{russakovsky2015imagenet}
O.~Russakovsky, J.~Deng, H.~Su, J.~Krause, S.~Satheesh, S.~Ma, Z.~Huang,
  A.~Karpathy, A.~Khosla, M.~Bernstein, et~al.
\newblock Imagenet large scale visual recognition challenge.
\newblock {\em IJCV}, 115(3):211--252, 2015.

\bibitem{shi2013bayesian}
Z.~Shi, T.~M. Hospedales, and T.~Xiang.
\newblock Bayesian joint topic modelling for weakly supervised object
  localisation.
\newblock In {\em ICCV}, pages 2984--2991, 2013.

\bibitem{Simonyan15}
K.~Simonyan and A.~Zisserman.
\newblock Very deep convolutional networks for large-scale image recognition.
\newblock In {\em ICLR}, 2015.

\bibitem{siva2012defence}
P.~Siva, C.~Russell, and T.~Xiang.
\newblock In defence of negative mining for annotating weakly labelled data.
\newblock In {\em ECCV}, pages 594--608. Springer, 2012.

\bibitem{song2014learning}
H.~O. Song, R.~B. Girshick, S.~Jegelka, J.~Mairal, Z.~Harchaoui, T.~Darrell,
  et~al.
\newblock On learning to localize objects with minimal supervision.
\newblock In {\em ICML}, pages 1611--1619, 2014.

\bibitem{song2014weakly}
H.~O. Song, Y.~J. Lee, S.~Jegelka, and T.~Darrell.
\newblock Weakly-supervised discovery of visual pattern configurations.
\newblock In {\em NIPS}, pages 1637--1645, 2014.

\bibitem{Tang_2016_CVPR}
Y.~Tang, J.~Wang, B.~Gao, E.~Dellandrea, R.~Gaizauskas, and L.~Chen.
\newblock Large scale semi-supervised object detection using visual and
  semantic knowledge transfer.
\newblock In {\em CVPR}, June 2016.

\bibitem{uijlings2013selective}
J.~R. Uijlings, K.~E. Van De~Sande, T.~Gevers, and A.~W. Smeulders.
\newblock Selective search for object recognition.
\newblock {\em IJCV}, 104(2):154--171, 2013.

\bibitem{wang2014weakly}
C.~Wang, W.~Ren, K.~Huang, and T.~Tan.
\newblock Weakly supervised object localization with latent category learning.
\newblock In {\em ECCV}, pages 431--445. Springer, 2014.

\bibitem{yang2016weakly}
K.~Yang, D.~Li, Y.~Dou, S.~Lv, and Q.~Wang.
\newblock Weakly supervised object detection using pseudo-strong labels.
\newblock {\em arXiv preprint arXiv:1607.04731}, 2016.

\bibitem{zitnick2014edge}
C.~L. Zitnick and P.~Doll{\'a}r.
\newblock Edge boxes: Locating object proposals from edges.
\newblock In {\em ECCV}, pages 391--405. Springer, 2014.

\end{thebibliography}
}

\end{document}